\documentclass[10pt,twocolumn,letterpaper]{article}

\usepackage{cvpr}              

\usepackage[dvipsnames]{xcolor}

\usepackage{graphicx}
\usepackage{amsmath}
\usepackage{amssymb}
\usepackage{booktabs}

\usepackage{wrapfig}
\usepackage{multirow}
\usepackage{pifont}
\usepackage{subfloat}

\definecolor{cvprblue}{rgb}{0.21,0.49,0.74}
\usepackage[pagebackref,breaklinks,colorlinks,citecolor=cvprblue]{hyperref}

\usepackage[capitalize]{cleveref}
\usepackage[accsupp]{axessibility}

\crefname{section}{Sec.}{Secs.}
\Crefname{section}{Section}{Sections}
\Crefname{table}{Table}{Tables}
\crefname{table}{Tab.}{Tabs.}

\def\paperID{1647} 
\def\confName{CVPR}
\def\confYear{2024}

\title{Vision-and-Language Navigation via Causal Learning}

\author{
Liuyi Wang,\,
Zongtao He,\,
Ronghao Dang,\,
Mengjiao Shen,\,
Chengju Liu\thanks{Corresponding author.},\,
Qijun Chen$^*$ \\
School of Electronic and Information Engineering, Tongji University, Shanghai, China\\
{\tt\small \{wly, xingchen327, dangronghao, liuchengju, qjchen\}@tongji.edu.cn,\, mojo.shum@outlook.com}
}

\begin{document}
\maketitle

\begin{abstract}
In the pursuit of robust and generalizable environment perception and language understanding, the ubiquitous challenge of dataset bias continues to plague vision-and-language navigation (VLN) agents, hindering their performance in unseen environments. This paper introduces the \textbf{g}eneralized cr\textbf{o}ss-modal c\textbf{a}usal \textbf{t}ransformer (GOAT), a pioneering solution rooted in the paradigm of causal inference. By delving into both observable and unobservable confounders within vision, language, and history, we propose the \textbf{b}ack-door and \textbf{f}ront-door \textbf{a}djustment \textbf{c}ausal \textbf{l}earning (BACL and FACL) modules to promote unbiased learning by comprehensively mitigating potential spurious correlations. Additionally, to capture global confounder features, we propose a \textbf{c}ross-modal \textbf{f}eature \textbf{p}ooling (CFP) module supervised by contrastive learning, which is also shown to be effective in improving cross-modal representations during pre-training. Extensive experiments across multiple VLN datasets (R2R, REVERIE, RxR, and SOON) underscore the superiority of our proposed method over previous state-of-the-art approaches. Code is available at \url{https://github.com/CrystalSixone/VLN-GOAT}.
\end{abstract}

\section{Introduction}
\label{sec:intro}
Effective environment perception, language understanding, and historical utilization are at the core of vision-and-language navigation (VLN)~\cite{anderson2018vision}. Despite significant progress, deploying VLN in the real world remains a huge challenge, primarily due to diversities and uncertainties in environments and instructions. A key hindrance is dataset bias~\cite{zhang2021diagnosing,zhu2022diagnosing}, \eg, agents may overfit to familiar visual environments, resulting in diminished performance in environments with diverse appearances and layouts~\cite{hu2019you}. This over-reliance on specific patterns, like biased structural trajectories and repeated entity components, raises concerns about the robustness and generalizability of VLN systems.
\begin{figure}[tb]
    \centering
    \includegraphics[width=0.98\linewidth]{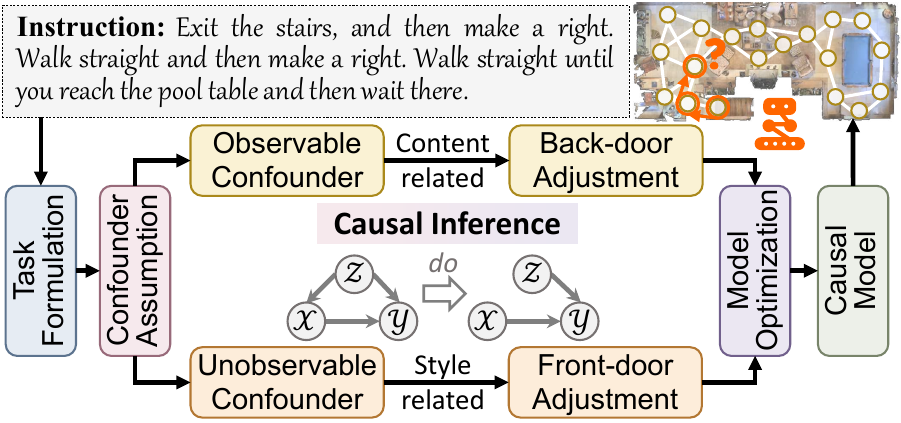}
    \caption{In response to language instructions, the VLN agent is required to navigate to the target based on visual cues. This paper introduces a causal learning pipeline using the \textit{do}-operator to reduce bias from confounders in VLN action prediction.}
    \label{fig:intro}
\end{figure}

One way to mitigate dataset bias in VLN is to build broader and more diverse datasets, which is what numerous recent studies have focused on. These include using speaker models to generate pseudo instructions~\cite{fried2018speaker,wang2022counterfactual,dou2022foam,wang2023pasts,wang2023res,wang2022less}, synthesizing cross-environmental trajectories~\cite{jain2019stay,liu2021vision}, transferring image styles~\cite{li2022envedit,li2023improving}, collecting data from the web~\cite{majumdar2020improving,guhur2021airbert,lin2023learning,wang2023scaling}, and labeling more fine-grained, entity-aligned instructions~\cite{he2023mlanet,cui2023grounded,ku2020room}. However, 
achieving a perfectly balanced dataset devoid of bias is nearly impossible. Consequently, we often find ourselves caught in a cycle of ``creating a dataset" - ``identifying bias" - ``creating a new dataset". Therefore, we are prompted to move from diagnosis to treatment~\cite{yang2021deconfounded}, transitioning from the continuous collection of new datasets to the development of unbiased models that can confront and mitigate bias.

However, existing methods that focus on model designing mainly concentrate on introducing more types of inputs (\eg, objects~\cite{wang2023dual,lin2022adapt,hong2023learning,an2021neighbor,dang2023search,dang2023multiple} and depth~\cite{an2022bevbert,liu2023bird,zhang20233d,he2023learning}), or constructing the global graph~\cite{chen2021history,chen2022think,nanwani2023instance} to represent environments. These efforts, although valuable, often overlook underlying dataset biases and the essential causal logic behind the task. In fact, the reason why humans can well execute various instructions and navigate in unknown environments is that we can learn the inherent causality of events beyond biased observation, achieving good analogical association capability. Therefore, for the first time, we propose to use causal inference~\cite{pearl2009causality} to equip VLN agents with similar cognitive abilities that we have, and then allow them to make more reasoned decisions.

Then, how to develop such a causal inference capability? Although there is no single answer, we propose to exploit the concepts of \textit{intervention}~\cite{pearl2018book} -- technique that uses a \textit{do}-operator to alleviate the negative effects raised by confounders. Here, confounders are variables that influence both inputs and outcomes, creating spurious correlations and biases. Intervention empowers researchers to mitigate the impact of confounders, enabling the model to grasp the causation of events during data fitting. However, given the fact that VLN is such a complex task that involves cross-modal inputs and a long-term decision-making process, it is challenging to identify underlying confounders and apply intervention to debias through network learning.

To address the above challenges, we propose a \textbf{g}eneralized cr\textbf{o}ss-modal c\textbf{a}usal \textbf{t}ransformer (GOAT) approach that enables the VLN model to alleviate the negative effects raised by confounders, thus achieving causal decision-making (\cref{fig:intro}). Firstly, we propose a unified structural causal model to describe the VLN system, involving two distinct categories of confounders: \textit{observable} and \textit{unobservable}. Observable confounders are content-related and easily identifiable (\eg, the keywords in instructions and the room references in environments), whereas unobservable confounders refer to intricate stylistic nuances that are harder to discern but can impact the overall system (\eg, decoration styles in vision, sentence patterns in language, and trajectory trends in history).
Then, we propose to address these confounders via two causal learning modules that are based on back-door and front-door adjustments~\cite{pearl2018book} (namely BACL and FACL), respectively. 
Furthermore, to build global dictionaries for representing confounders, we devise a cross-modal feature pooling (CFP) module to effectively aggregate long-sequential features. 
Contrastive learning~\cite{radford2021learning} is adopted to optimize CFP, serving as an additional auxiliary task during pre-training.
As demonstrated by thorough experiments, our findings reveal the impact of integrating causal learning to deconfound biases on cross-modal inputs, offering valuable insights for enhancing generalization in similar tasks across diverse scenarios.

To summarize, our main contributions are as follows:

\begin{itemize}
\setlength{\itemsep}{0pt}
\setlength{\parsep}{0pt}
\setlength{\parskip}{0pt}
    \item We propose a unified structural causal model for VLN by comprehensively considering the observable and unobservable confounders hidden in different modalities.
    \item we propose BACL and FACL, using the back-door and front-door adjustments to allow end-to-end unbiased cross-modal intervention and decision-making.
    \item We propose CFP, a cross-modal feature pooling module designed to aggregate sequence features for semantic alignment and confounder dictionaries construction.
    \item Our GOAT model demonstrates exceptional generalization across multiple VLN datasets (R2R~\cite{anderson2018vision}, RxR~\cite{ku2020room}, REVERIE~\cite{qi2020reverie}, and SOON~\cite{Zhu_2021_SOON}), outperforming existing state-of-the-art methods. A comprehensive causal learning pipeline is presented to inspire future research.
\end{itemize}

\section{Related Work}
\label{sec:related work}

\textbf{Vision-and-Language Navigation (VLN)} requires agents to navigate to specific locations~\cite{anderson2018vision,ku2020room} or find target objects~\cite{qi2020reverie,Zhu_2021_SOON} in real visual environments based on natural instructions. Its practicability has led to significant interest, showing potential in fundamental embodied AI skills. Initial models relied on recurrent neural networks~\cite{wang2020vision,an2021neighbor,dang2022unbiased,he2023mlanet}. Transformer-based models~\cite{hao2020towards,hong2021vln,chen2021history,chen2022think,wang2023dual} brought substantial progress due to their powerful long-distance encoding. However, small-scale datasets in VLN were found to cause bias, leading to serious overfitting. Consequently, several approaches were devised to tackle this challenge. Speaker-follower frameworks~\cite{fried2018speaker,tan2019learning,wang2023pasts,wang2023res,kamath2023new} used the speaker model to generate pseudo instructions. VLN-BERT~\cite{majumdar2020improving}, AirBERT~\cite{guhur2021airbert}, and Lily~\cite{lin2023learning} collected trajectory-instruction pairs from diverse sources. REM~\cite{liu2021vision} and EnvEdit~\cite{li2022envedit} created new environments by editing existing environments. Although these methods improve generalization, they cannot completely eliminate inherent dataset biases. Therefore, we propose to develop an unbiased model by equipping the agent with the causal inference capability to learn cause-effect relations during data fitting, enabling them to adapt adeptly to diverse situations. 

\textbf{Causal Inference} is an emerging technique exploring task causality~\cite{pearl2018book}, leading to a surge in efforts to integrate it with deep learning, in tasks like image recognition~\cite{wang2021causal,zhang2022multiple,wang2023meta}, image captioning~\cite{yang2021deconfounded,liu2022show}, and visual question answering~\cite{niu2021counterfactual,li2022representation}. One popular way is to use the adjustment technique to alleviate the negative effects caused by confounders, and some studies exploring the use of counterfactuals~\cite{niu2021counterfactual,abbasnejad2020counterfactual,parvaneh2020counterfactual}. This paper emphasizes the adjustment method due to its practicality. However, most of the existing causal learning tasks are simple without considering more challenging tasks like VLN. Additionally, current methods apply back-door~\cite{wang2020visual,yue2020interventional,zhang2020devlbert,liu2022show} or front-door~\cite{yang2021deconfounded,yang2021causal,liu2023cross} adjustments separately across modalities, lacking comprehensive confounder assumptions and complete bias corrections. In this paper, we propose to simultaneously tackle both observable and unobservable confounders in vision, language, and history. This approach significantly reduces overall bias, enhancing the generalization capabilities of embodied VLN agents.

\section{Preliminary}
\label{sec:pre}

\subsection{Task Formulation}
\label{subsec: formulation}
The VLN task~\cite{anderson2018vision} involves an embodied agent following natural language instructions to navigate real indoor environments. Matterport3D simulator~\cite{chang2017matterport3d} is used to allow interaction, where the environment is provided as graphs with connected navigable nodes. The agent receives natural language instructions $\mathcal{I}=\{w_1,w_2,...,w_L\}$ with $L$ words, and the current panorama separated into 36 sub-images $\mathcal{V}=\{v_1,v_2,...,v_{36}\}$~\cite{fried2018speaker}. The agent also knows its current heading $\theta$ and elevation $\phi$.
During navigation, the agent needs to select the next point from nearby candidates or predicts the \texttt{stop} signal based on visual cues. Success is defined when the stop location is within 3 meters of the ground-truth position. For the goal-oriented task, REVERIE~\cite{qi2020reverie} and SOON~\cite{Zhu_2021_SOON} additionally require locating the target object at the final destination.

\subsection{Structural Causal Model of VLN}
\label{subsec: causal graph}
\begin{figure}
    \centering
    \includegraphics[width=0.8\linewidth]{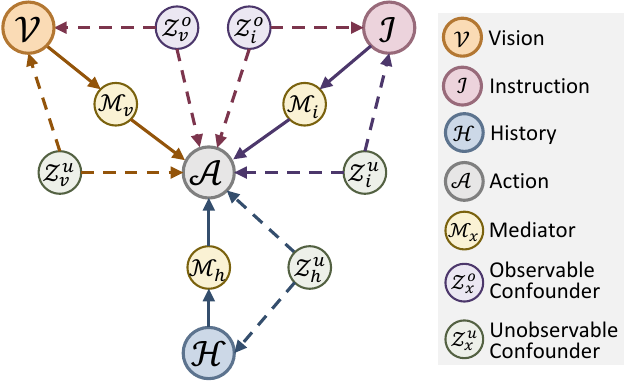}
    \caption{Illustration of the structural causal model of VLN.}
    \label{fig:SCM}
\end{figure}
\begin{figure*}[thb]
    \centering
    \includegraphics[width=0.98\linewidth]{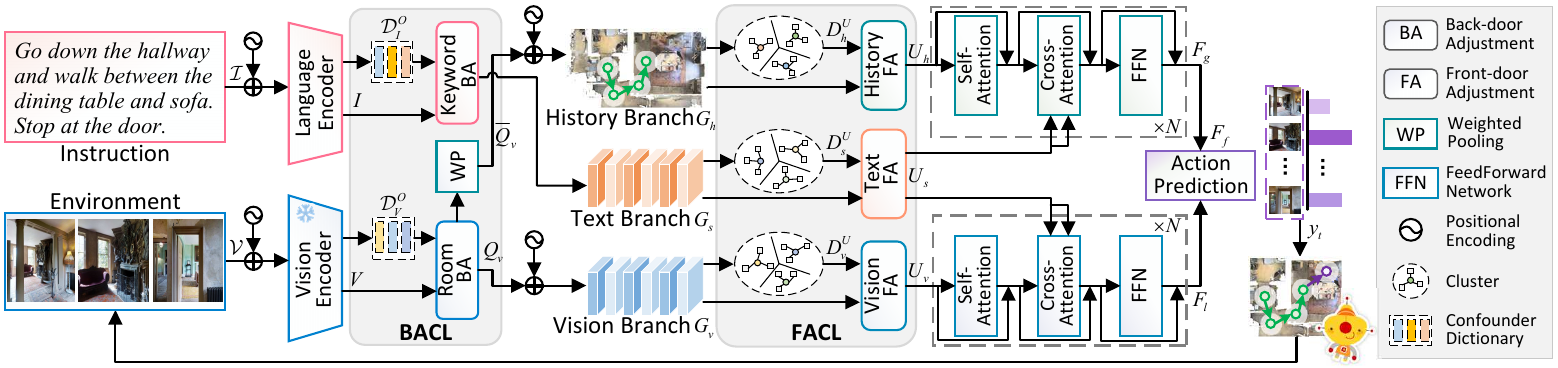}
    \caption{Framework of GOAT, built on the foundation of the dual-scale graph transformer~\cite{chen2022think}. Back-door and front-door adjustment causal learning mechanisms are used for mitigating spurious correlations, enabling unbiased feature learning and decision-making.}   
    \label{fig:overview}
\end{figure*}

As illustrated in~\cref{fig:SCM}, we construct a structural causal model capturing the relationships among the key variables in VLN: visual observation $\mathcal{V}$, linguistic instruction $\mathcal{I}$, decision history $\mathcal{H}$, and action prediction $\mathcal{A}$. To clarify, we use $\mathcal{X}$ to denote inputs ($\mathcal{X}=\{\mathcal{V},\mathcal{I},\mathcal{H}\}$) and $\mathcal{Y}$ for output ($\mathcal{Y}=\mathcal{A}$). In this directed acyclic graph, the starting and ending points represent the cause and effect, respectively. Traditional VLN methods focus on learning the observational association $P(\mathcal{Y}|\mathcal{X})$, overlooking the ambiguity introduced by confounders $\mathcal{Z}$ in the back-door path $\mathcal{X} \leftarrow \mathcal{Z} \rightarrow \mathcal{Y}$. Here, confounders are extraneous variables that influence both causes and effects, e.g., frequently occurring content or specific attributes.
$\mathcal{Z} \rightarrow \mathcal{X}$ arises since the combined probability of input data is inevitably affected by the limited resources available in the real world when collection and simulation. Additionally, $\mathcal{Z} \rightarrow \mathcal{Y}$ exists because collected environments, labeled instructions, or sampled trajectories also affect the probability of action distributions. These confounder links enable spurious shortcuts during training but can be detrimental in new situations.
We propose to distinguish hidden confounders from different modalities into \textit{observable} and \textit{unobservable} categories, enhancing our prior knowledge integration and the rationality of assumption. Concretely, observable confounders encompass instances that can be recognized (\eg, room references $z^o_v$ and guiding keywords $z^o_i$). In contrast, unobservable confounders consist of intricate patterns and style-related elements that are challenging to qualitatively describe (\eg, decoration style in vision $z^u_v$, sentence pattern in language $z^u_i$, and trajectory trend in history $z^u_h$). Since we cannot explicitly model unobservable confounders $\mathcal{Z}^u$, 
the additional mediators $\mathcal{M}$ are inserted between $\mathcal{X}$ and $\mathcal{Y}$ to establish front-door paths $\mathcal{X} \rightarrow \mathcal{M} \rightarrow \mathcal{Y}$. Detailed adjustment methods are introduced in subsequent sections.

\section{Methodology}
\label{sec:method}
The overview of the GOAT model is shown in~\cref{fig:overview}. 
The proposed back-door and front-door adjustment causal learning modules are detailed in~\cref{subsec:Back-door} and~\cref{subsec:frontdoor}, respectively. The cross-modal feature pooling method is subsequently introduced in~\cref{subsec:TIM}. Finally, a practical causal learning pipeline is presented in~\cref{subsec:pipeline}.

\subsection{Observable Causal Inference}
\label{subsec:Back-door}
\begin{figure}[]
    \centering
    \includegraphics[width=1.0\linewidth]{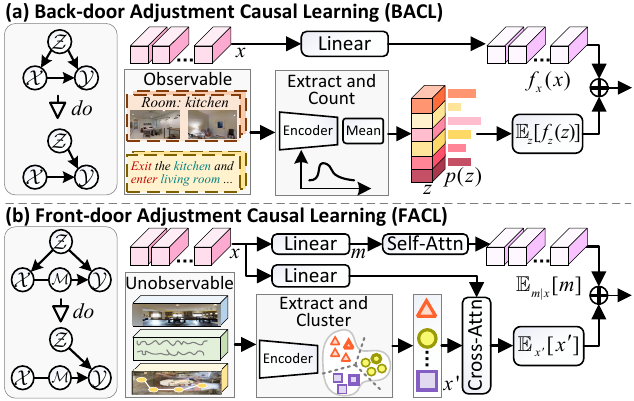}
    \caption{Illustration of BACL and FACL.}
    \label{fig:back_front}
\end{figure}
\noindent\textbf{Back-door Adjustment Causal Learning (BACL).}
Based on Bayes's theorem, the typical observational likelihood is as $P(\mathcal{Y}|\mathcal{X})=\sum_{z}P(\mathcal{Y}|\mathcal{X},z)\underline{P(z|\mathcal{X})}$, where $P(z|\mathcal{X})$ could bring biased weights. \textit{Do}-operator~\cite{pearl2018book} provides scientifically sound methods for determining causal effects by severing the back-door link between $\mathcal{Z}$ and $\mathcal{X}$. According to the invariance and independence rules~\cite{glymour2016causal}, we have:
{\setlength\abovedisplayskip{1pt}
\setlength\belowdisplayskip{1pt}
\begin{align}
    P(\mathcal{Y}|\textit{do}(\mathcal{X})) &= \sum_{z} P(\mathcal{Y}|\textit{do}(\mathcal{X}),z)P(z|\textit{do}(\mathcal{X})) \\ 
    &= \sum_{z} P(\mathcal{Y}|\mathcal{X},z)\underline{P(z)}\label{eq:back-door}
\end{align}}

In such case, the intervention is achieved by blocking the back-door path $\mathcal{Z} \rightarrow \mathcal{X}$, making $\mathcal{X}$ have a fair opportunity to incorporate causality-related factors for prediction. Previous methods~\cite{zhang2020devlbert,yang2021deconfounded,liu2023cross} used the NWGM approximation~\cite{baldi2014dropout} to directly pursue causal learning to the final outputs. However, these methods limit the intervention only to the network's final Softmax layer, overlooking possible biased features in shallow layers. Since the conditional probability is implicit in pattern recognition made by the trained neural network~\cite{nie2018deep}, we release the target effect of causal hypothesis to \textit{learned features} rather than merely \textit{outputs}.
Obtaining unbiased features leads to unbiased predictions. 
Consequently, the specific network module is formulated as $f(\boldsymbol{x},\boldsymbol{z})$, and the implementation of~\cref{eq:back-door} becomes:
{\setlength\abovedisplayskip{1pt}
\setlength\belowdisplayskip{2pt}
\begin{align}
\mathcal{B}(\boldsymbol{x},\boldsymbol{z})=\mathbb{E}_z[f(\boldsymbol{x},\boldsymbol{z})] \label{eq:back_door_exp}
\end{align}}

The linear function is used as $f(\boldsymbol{x},\boldsymbol{z})=f_x(\boldsymbol{x})+f_z(\boldsymbol{z})$. Then~\cref{eq:back_door_exp} becomes $f_x(\boldsymbol{x})+\mathbb{E}_z[f_z(\boldsymbol{z})]$. To obtain $\mathbb{E}_z[f_z(\boldsymbol{z})]$, there are two prevalent approaches: statistic-based~\cite{wang2020visual,liu2023cross} and attention-based methods~\cite{yang2021deconfounded,liu2022show}:
{\setlength\abovedisplayskip{2pt}
\setlength\belowdisplayskip{2pt}
\begin{align}
    &\textbf{Statistic}: \mathbb{E}_z[f_z(\boldsymbol{z})]=\sum_i \frac{|z_i|}{\sum_j|z_j|} f_z(\boldsymbol{z}_i) \\
    &\textbf{Attention}: \mathbb{E}_z[f_z(\boldsymbol{z})]=\sum_i \frac{\text{exp}(\boldsymbol{h}\boldsymbol{z}_i^{T})}{\sum_j \text{exp}({\boldsymbol{h} \boldsymbol{z}_j^{T})}} f_z(\boldsymbol{z}_i)
\end{align}}
where $|z_i|$ denotes the number of $z$ belonging to the $i$-th category in the training set, and $\boldsymbol{h}$ means hidden features. The illustration is shown in~\cref{fig:back_front}(a).
Our experiments in~\cref{subsec:ablation} reveal modality-specific calculation preferences.

\noindent\textbf{BACL in Text Content.} In VLN instructions like ``Exit the office and turn right into the kitchen," essential guiding elements such as directions (\eg, ``exit" and ``right") and landmarks (\eg, ``office" and ``kitchen") play significant roles. These keywords which are common causes of instruction construction and action distribution, serve as observable confounders. Firstly, we build the text keyword dictionary $\mathcal{D}^O_I=[\boldsymbol{z}^o_{i,1},\boldsymbol{z}^o_{i,2},...,\boldsymbol{z}^o_{i,K}]$ with $K$ classes to store confounder features. Direction-and-landmark keywords are extracted based on their part-of-speech tags~\cite{wang2023dual}. We use the pre-trained RoBERTa~\cite{liu2019roberta} to obtain feature representations for each extracted token $\boldsymbol{f}^{o}_{i}$. Since the same word can have different features across sentences, we calculate the average feature for each keyword: $\boldsymbol{z}^o_{i,n}=\frac{1}{|z^o_{i,n}|}\sum_j\boldsymbol{f}^{o}_{i,n,j}$. Subsequently, the text content causal representation $G_s$ is calculated as follows:
{\setlength\abovedisplayskip{1pt}
\setlength\belowdisplayskip{1pt}
\begin{gather}
    I=\text{RoBERTa}(\psi_t(\mathcal{I})+\psi_t(\mathcal{P})),\, Z_k=\text{LN}(\phi_k(\mathcal{D}^O_I)) \\
    G_s = \text{LN}[\phi_i(\mathcal{B}(I,Z_k))]
\end{gather}}
where $\psi(\cdot)$ and $\phi(\cdot)$ denote the learnable embedding layer and the full-connection layer, respectively. The absolute positional encoding $\mathcal{P}$~\cite{vaswani2017attention} is added to present the position information, and the layer normalization $\text{LN}$~\cite{Ba2016LayerN} is employed for stabilizing hidden states during training. 

\noindent\textbf{BACL in Vision Content.} For each step, the panorama $\mathcal{V}$ is divided into 36 sub-images. Since existing VLN datasets primarily involve indoor room navigation, visual room references are treated as observable confounders. CLIP~\cite{radford2021learning} is used to extract image features. Since room labels are not directly provided, we employ BLIP~\cite{li2022blip}, a pre-trained VQA model to capture the room information for each image, by querying the model with a fixed prompt \textit{``what kind of room is this?"} The average value of each room reference type is calculated, forming a visual room reference dictionary $\mathcal{D}^{O}_{V}=[\boldsymbol{z}^o_{v,1},\boldsymbol{z}^o_{v,2},...,\boldsymbol{z}^o_{v,M}]$, where $M$ is the number of room types. Additionally, the matrix $\gamma = [(\sin{\theta_i},\cos{\theta_i},\sin{\eta_i},\cos{\eta_i})_{i=1}^{36}]$ is used to present the direction of each image's shift relative to the agent, where $\theta$ and $\eta$ denote the heading and elevation direction. If there are additional object features (for goal-oriented tasks), they are concatenated with image features. Subsequently, a 2-layer transformer encoder is used to capture spatial dependencies. The above process is formulated as follows:
{\setlength\abovedisplayskip{1pt}
\setlength\belowdisplayskip{1pt}
\begin{gather}
    V=\text{CLIP}(\mathcal{V}),\, Z_r=\text{LN}(\phi_r(\mathcal{D}^O_V)) \\
    V_v = \text{LN}[\phi_v(\mathcal{B}(V,Z_r))]\\
    Q_v = \text{Trans}(V_v+\psi_d(\gamma))
\end{gather}}

\subsection{Unobservable Causal Inference}
\label{subsec:frontdoor}

\noindent\textbf{Front-door Adjustment Causal Learning (FACL).} 
In the previous section, we employed the back-door adjustment technique to handle bias caused by observable confounders. However, there are additional unobservable confounders that cannot be explicitly captured and modeled in advance. 
To address this, we introduce another technique - front-door adjustment~\cite{glymour2016causal}. 
As shown in~\cref{fig:back_front}(b), an additional mediator $\mathcal{M}$ is inserted between inputs and outcomes to construct the front door path $\mathcal{X}\rightarrow \mathcal{M} \rightarrow \mathcal{Y}$. In VLN, an attention-based model $P(\mathcal{Y}|\mathcal{X})=\sum_m P(\mathcal{Y}|m)P(m|\mathcal{X})$ will select key regions $\mathcal{M}$ from inputs $\mathcal{X}$ for action prediction $\mathcal{Y}$. Therefore, the model inference can be represented by two parts: the feature selector $\mathcal{X}\rightarrow\mathcal{M}$ which selects suitable knowledge $\mathcal{M}$ from $\mathcal{X}$, and the action predictor $\mathcal{M}\rightarrow\mathcal{Y}$ which exploits $\mathcal{M}$ to predict $\mathcal{Y}$. To eliminate spurious correlation brought by unobservable confounder $\mathcal{Z}$, we simultaneously deploy \textit{do}-operator to $\mathcal{X}$ and $\mathcal{M}$:
{\setlength\abovedisplayskip{1pt}
\setlength\belowdisplayskip{1pt}
\begin{align}
    P(\mathcal{Y}|\textit{do}(\mathcal{X})) &= \sum_{m} P(\mathcal{Y}|\textit{do}(m))P(m|\textit{do}(\mathcal{X})) \\
    &= \sum_{x'}P(x') \sum_{m} P(\mathcal{Y}|m,x')P(m|\mathcal{X}) \\
    &= \mathbb{E}_{x'} \mathbb{E}_{m|x}[P(\mathcal{Y}|x',m)] \label{eq:front-door}
\end{align}}
where $x'$ denotes potential input samples of the whole representation space, different from current inputs $\mathcal{X}=x$. We use the bold symbol $\boldsymbol{m}$ to denote the in-sampling features obtained by the feature extractor acting on the current input, and $\boldsymbol{x}'$ to mean the cross-sampling features randomly sampled by the K-means-based feature selector from the entire training samples. Based on the linear mapping model, ~\cref{eq:front-door} becomes $\mathbb{E}_{m|x}[\boldsymbol{m}]+\mathbb{E}_{x'}[\boldsymbol{x}']$. 
As it is intractable to get a closed-form solution of expectations involving the complex representation space, the estimation is achieved by the query mechanism. Two embedding functions~\cite{yang2021deconfounded,liu2023cross} are used to transmit input $\boldsymbol{x}$ into two query sets $\boldsymbol{g}_1=q_1(\boldsymbol{x})$ and $\boldsymbol{g}_2=q_2(\boldsymbol{x})$. Then, the front-door adjustment is approximated as follows:
{\setlength\abovedisplayskip{0pt}
\setlength\belowdisplayskip{0pt}
\small
\begin{gather}
\mathbb{E}_{x'}[\boldsymbol{x}']\approx\sum_{x'} P(\boldsymbol{x}'|\boldsymbol{g}_1)\boldsymbol{x}'= \sum_i\frac{\text{exp}(\boldsymbol{g}_1 \boldsymbol{x}'^T_i)}{\sum_{j}\text{exp}(\boldsymbol{g}_1 \boldsymbol{x}'^T_j)} \boldsymbol{x}_i' \\
\mathbb{E}_{m|x}[\boldsymbol{m}]\approx\sum_{m} P(\boldsymbol{m}|\boldsymbol{g}_2)\boldsymbol{m}= \sum_i\frac{\text{exp}(\boldsymbol{g}_2 \boldsymbol{m}_i^T)}{\sum_{j}\text{exp}(\boldsymbol{g}_2 \boldsymbol{m}^T_j)} \boldsymbol{m}_i \\
\mathcal{F}(\boldsymbol{x},\boldsymbol{x}') = \mathbb{E}_{x'}[\boldsymbol{x}'] + \mathbb{E}_{m|x}[\boldsymbol{m}] \label{eq:FA}
\end{gather}}
The above process can be efficiently implemented using the multi-head attention~\cite{vaswani2017attention}, enabling seamless integration of causal adjustments into existing transformer-based frameworks with minimal modifications.

\noindent\textbf{FACL in Text, Vision, and History.}
Considering the characteristics of VLN, we propose to eliminate unobservable confounders from three kinds of inputs in VLN, \ie, vision, language, and history. 
First, following previous graph-based methods~\cite{wang2023dual,chen2022think}, we construct the vision sequence $G_v=\{\texttt{[STOP]},\texttt{[MEM]},Q_v\}$ and the history sequence $G_h=\{\texttt{[STOP]},\texttt{[MEM]},\{\bar{Q}_t\}_{t=1}^{T}\}$ by adding the additional token for presenting stop and recurrent memory states, respectively. $\bar{Q}_t$ means the learned weight sum of panoramic features for the $t$-th step. To condense the lengthy sequence of features and generate global features $\boldsymbol{x}'$ for cross-sampling, we devise the CFP module (as described in~\cref{subsec:TIM}) with the attentive pooling mechanism to construct confounder dictionaries for vision, history, and instruction, denoted as $D^U_v,D^U_h$, and $D^U_s$, respectively. Then the causality-enhanced features $R_v,R_h$ and $R_s$ are calculated based on~\cref{eq:FA}:
{\setlength\abovedisplayskip{0pt}
\setlength\belowdisplayskip{0pt}
\small
\begin{gather}
    R_v=\mathcal{F}(G_v,D^U_v), R_h=\mathcal{F}(G_h,D^U_h), R_s=\mathcal{F}(G_s,D^U_s)
\end{gather}}
Furthermore, we introduce an adaptive gate fusion (AGF) method to enhance the stability of learning by integrating causality-enhanced features with the original context features for each modality:
{\setlength\abovedisplayskip{1pt}
\setlength\belowdisplayskip{1pt}
\begin{gather}
    \omega_x = \delta(R_xW_x+G_xW_g+b) \\
    U_x = \omega_x \odot R_x + (1-\omega_x)\odot G_x
\end{gather}}
where $\delta$ and $\odot$ mean the Sigmoid function and element-wise multiplication. Suppose $R_x$ and $G_x \in \mathbb{R}^{L_x\times d_h}$, then $W_{x/g}\in \mathbb{R}^{d_h\times 1}$ and $b\in \mathbb{R}^{L_x\times 1}$ are learnable parameters. Next, the cross-modal fused local features $F_l$ and global features $F_g$ are obtained by the cross-attention encoders $\mathcal{C}$ from METER~\cite{dou2022empirical}. The dynamic fusion $\mathcal{DF}$~\cite{chen2022think} followed by Softmax $\mathcal{SF}$ is applied for action prediction:
{\setlength\abovedisplayskip{1pt}
\setlength\belowdisplayskip{1pt}
\begin{gather}
    F_l=\mathcal{C}(U_v,U_s,U_s),\,F_g=\mathcal{C}(U_h,U_s,U_s)\\
    F_f=\mathcal{DF}(F_l,F_g),\,y_t=\mathcal{SF}(F_f)
\end{gather}}

The cross-entropy loss is used to optimize the network:
{\setlength\abovedisplayskip{1pt}
\setlength\belowdisplayskip{1pt}
\begin{gather}
    \mathcal{L}_{ce}=\sum_{t=1}^{T}-\log P(y_t^*|\mathcal{I},\mathcal{V}_t,\mathcal{H}_{1:t-1})
\end{gather}}

\subsection{Cross-modal Feature Pooling}
\label{subsec:TIM}
\begin{figure}[tb]
    \centering
    \includegraphics[width=1.0\linewidth]{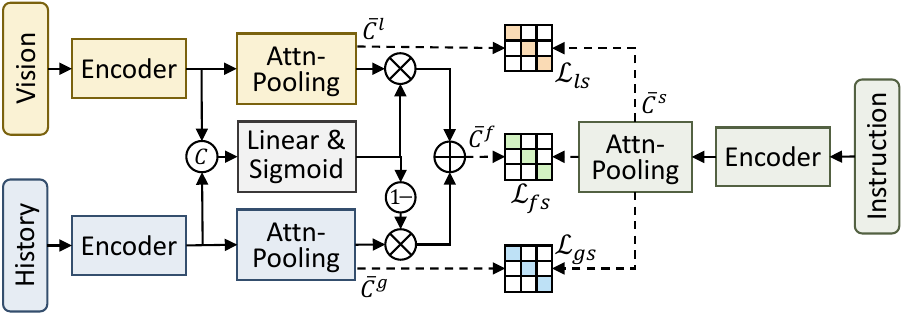}
    \caption{Illustration of the cross-modal feature pooling (CFP).}
    \label{fig:tim}
\end{figure}
One challenge of implementing the front-door adjustment in VLN is constructing efficient dictionaries for global features from long sequences. This requires compressing sequential features of varying lengths into a unified feature space to represent each sample effectively. 
Let $H\in \mathbb{R}^{L\times d_h}$ be the sequential features, the following attentive pooling is used to effectively compress the sequence length:
{\setlength\abovedisplayskip{2pt}
\setlength\belowdisplayskip{2pt}
\begin{gather}
    A=\mathcal{T}(H),\, \alpha=\mathcal{SF}(AW_a),\, \bar{H}=\mathcal{T}(\alpha^TH)
\end{gather}}
where $\mathcal{T}$ denote the Tanh activation, $W_a \in \mathbb{R}^{d_h\times 1}$ is the learnable attention matrix, and $\bar{H}\in \mathbb{R}^{1\times d_h}$. As shown in~\cref{fig:tim}, for vision, history, local-global fusion, and text features, we use one transformer layer as the encoder followed by the attentive pooling to obtain flattened features $\bar{C}^l,\bar{C}^g,\bar{C}^f$, and $\bar{C}^s$, respectively. Then, we adopt contrastive learning~\cite{shen2021much,lin2022adapt,wang2023res} to optimize this cross-modal feature pooling (CFP) module, meanwhile improving semantic alignments for different modalities. The contrastive loss $\mathcal{L}_{ls}$ is constructed as:
\begin{equation}
    \label{eq_contrastive}
    \begin{aligned}
        \mathcal{L}_{ls}=&-\frac{1}{2B}\sum_{j=1}^B\log \frac{\exp (\langle \bar{C}^l_j,\bar{C}^s_j\rangle /t)}{\sum_{k=1}^{B}\exp(\langle \bar{C}^l_j,\bar{C}^s_k \rangle /t)} \\
        &-\frac{1}{2B}\sum_{k=1}^B\log \frac{\exp (\langle \bar{C}^l_k,\bar{C}^s_k \rangle/t)}{\sum_{j=1}^{B}\exp(\langle \bar{C}^l_j,\bar{C}^s_k \rangle/t)}
    \end{aligned}
\end{equation}
where $B$ and $t$ mean the batch size and temperature, respectively. Similarly, contrastive losses $\mathcal{L}_{gs}$ and $\mathcal{L}_{fs}$ are calculated by replacing $\bar{C}^l$ with $\bar{C}^g$ and $\bar{C}^f$. The overall CFP loss is the sum of these losses $\mathcal{L}_{\text{CFP}}=\mathcal{L}_{ls}+\mathcal{L}_{gs}+\mathcal{L}_{fs}$.

To enable the network more adaptive to characteristics for VLN and thus facilitate the building of the front-door confounder dictionaries for samples, we train the CFP module alongside other auxiliary tasks~\cite{chen2022think} during pre-training. Subsequently, the trained attentive pooling modules are used to extract global features for different modalities. In the fine-tuning stage, we employ the BACL and FACL with established dictionaries for intervention. The CFP offers dual benefits: it aligns different modalities more effectively during pre-training and provides a systematic approach for extracting coherent representations from sequence inputs.

\subsection{Causal Learning Pipeline}
\label{subsec:pipeline}
\begin{wrapfigure}{r}{0em}
    \centering
    \includegraphics[width=0.4\linewidth]{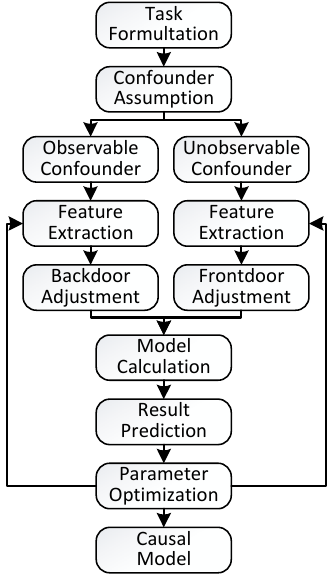}
    \caption{Pipeline of the causal learning.}
    \label{fig:pipeline}
\end{wrapfigure}
As shown in~\cref{fig:pipeline}, we summarize a causal learning pipeline that serves as a blueprint for similar learning-based methods. First, it begins with task formulation, where the specific task and its objectives are precisely defined. Next, the observable and unobservable confounders are explicitly assumed. Both back-door and front-door adjustment strategies are employed to tackle these confounders, either simultaneously or sequentially, contingent on task specifics. Then, the pipeline proceeds to model calculation and result prediction. Throughout network optimization, both network parameters and confounder features are continuously updated. Ultimately, this iterative process leads us to the development of a robust causal model capable of generating unbiased features, thereby advancing the generalizability of AI systems.

\begin{table*}[t]
\centering
\renewcommand{\arraystretch}{0.85}
\small
\begin{tabular}{l|cccc|cccc|cccc}
\toprule
\multirow{2}{*}{Method} & \multicolumn{4}{c|}{Validation Seen} & \multicolumn{4}{c|}{Validation Unseen} & \multicolumn{4}{c}{Test Unseen} \\
 & SR$\uparrow$ & SPL$\uparrow$ & NE$\downarrow$ & OSR$\uparrow$ & SR$\uparrow$ & SPL$\uparrow$ & NE$\downarrow$ & OSR$\uparrow$ & SR$\uparrow$ & SPL$\uparrow$ & NE$\downarrow$ & OSR$\uparrow$ \\ \midrule
HAMT~\cite{chen2021history}& 76 & 72 & 2.51 & 82 & 66 & 61 & 3.29 & 73 & 65 & 60 & 3.93 & 72 \\
DUET~\cite{chen2022think} & 79 & 73 & 2.28 & 86 & 72 & 60 & 3.31 & 81 & 69 & 59 & 3.65 & 76 \\
TD-STP~\cite{zhao2022target} & 77 & 73 & 2.34 & 83 & 70 & 63 & 3.22 & 76 & 67 & 61 & 3.73 & 72 \\
GeoVLN~\cite{huo2023geovln} & 79 & 76 & 2.22 & -- & 68 & 63 & 3.35 & -- & 65 & 61 & 3.95 & -- \\
DSRG~\cite{wang2023dual} & 81 & 76 & 2.23 & 88 & 73 & 62 & 3.00 & 81 & 72 & 61 & 3.33 & 78 \\
GridMM~\cite{wang2023gridmm} & -- & -- & -- & -- & 75 & 64 & 2.83 & -- & 73 & 62 & 3.35 & -- \\ 
GELA~\cite{cui2023grounded} & 76 & 73 & 2.39 & -- & 71 & 65 & 3.11 & -- & 67 & 62 & 3.59 & -- \\
EnvEdit~\cite{li2022envedit} & 77 & 74 & 2.32 & -- & 69 & 64 & 3.24 & -- & 68 & 64 & 3.59 & -- \\
BEVBert~\cite{an2022bevbert} & 81 & 74 & 2.17 & 88 & 75 & 64 & 2.81 & 84 & 73 & 62 & 3.13 & \textbf{81} \\ \midrule
\textbf{GOAT (Ours)} & \textbf{83.74} & \textbf{79.48} & \textbf{1.79} & \textbf{88.64} &\textbf{77.82}	&\textbf{68.13}&\textbf{2.40}&\textbf{84.72}&\textbf{74.57}&\textbf{64.94}&\textbf{3.04}&80.35 \\ \bottomrule
\end{tabular}
\caption{Comparison with other state-of-the-art methods on the R2R dataset~\cite{anderson2018vision}. `--': unavailable statistics.}
\label{tab:r2r}
\end{table*}

\section{Experiments}
\label{sec:experiments}

\subsection{Experimental Settings}
\noindent\textbf{1) Datasets.} 
We verify GOAT on two kinds of VLN benchmarks: fine-grained datasets (R2R~\cite{anderson2018vision} and RxR-English~\cite{ku2020room}), which provide long step-by-step navigation instructions, and goal-oriented datasets (REVERIE~\cite{qi2020reverie} and SOON~\cite{Zhu_2021_SOON}), which additionally requires for the target object. Formally, the datasets are partitioned into four splits: training, validation seen (sharing the same environments with the training set), validation unseen (having different environments from the training set), and test unseen sets (reported by the online leaderboard for fair comparison).

\noindent\textbf{2) Evaluation Metrics.}
In R2R, key metrics include Navigation Error (NE), Success Rate (SR), Oracle SR (OSR), and SR Weighted by Path Length (SPL). RxR adds Normalized Dynamic Time Warping (nDTW) and SR Weighted by Dynamic Time Warping (sDTW). REVERIE and SOON introduce Remote Grounding Success Rate (RGS) and RGS Weighted by Path Length (RGSPL). 

\noindent\textbf{3) Implementation Details.}
Our model consists of 6 transformer layers for text, 2 for panorama, and 3 for cross-modal encoding. We use CLIP-B/16~\cite{radford2021learning} for image feature extraction and initialize network weights with METER~\cite{dou2022empirical}. In pre-training, MLM~\cite{devlin2018bert}, SAP~\cite{chen2021history}, and the proposed CFP are for R2R and RxR. OG~\cite{Lin_2021_CVPR} is added for REVERIE and SOON. EnvEdit~\cite{li2022envedit} is employed for feature augmentation. The synthetic extended datasets~\cite{hao2020towards, wang2023res, wang2022less} are used for R2R, REVERIE, and RxR, respectively. Pre-training is done on a single Tesla V100 GPU for a maximum of 300K iterations by the AdamW~\cite{loshchilov2018decoupled} optimizer, with a batch size of 48 and a learning rate of $5\times 10^{-5}$. The numbers of classes of keywords and rooms are 74 and 50, and the temperature $t$ is set to 1. In fine-tuning, the front-door dictionaries are randomly sampled from the K-Means clustering features. As the text transformer RoBERTa~\cite{liu2019roberta} is involved in end-to-end training, the textual keywords dictionary is also iteratively updated. 
Speaker models with the environmental dropout~\cite{tan2019learning,wang2023res,wang2023pasts} are used to provide dynamic pseudo labels. We employ batch size 12 for R2R, REVERIE, and 5 for RxR and SOON, with a learning rate of $2\times 10^{-5}$ and a maximum of 100K iterations.

\begin{table*}[t]
\centering
\renewcommand{\arraystretch}{0.85}
\small
\begin{tabular}{@{}l|cccc|cccc|cccc@{}}
\toprule
\multirow{2}{*}{Method} & \multicolumn{4}{c|}{Validation Seen} & \multicolumn{4}{c|}{Validation Unseen} & \multicolumn{4}{c}{Test Unseen} \\
 & SR$\uparrow$ & SPL$\uparrow$ & RGS$\uparrow$ & RGSPL$\uparrow$ & SR$\uparrow$ & SPL$\uparrow$ & RGS$\uparrow$ & RGSPL$\uparrow$ & SR$\uparrow$ & SPL$\uparrow$ & RGS$\uparrow$ & RGSPL$\uparrow$ \\ \midrule
HAMT~\cite{chen2021history} & 43.29 & 40.19 & 27.20 & 25.18 & 32.95 & 30.20 & 18.92 & 17.28 & 30.40 & 26.67 & 14.88 & 13.08 \\
HOP+~\cite{qiao2023hop_plus} & 55.87 & 49.55 & 40.76 & 36.22 & 36.07 & 31.13 & 22.49 & 19.33 & 33.82 & 28.24 & 20.20 & 16.86 \\
DUET~\cite{chen2022think} & 71.75 & 63.94 & 57.41 & 51.14 & 46.98 & 33.73 & 32.15 & 23.03 & 52.51 & 36.06 & 31.88 & 22.06 \\
DSRG~\cite{wang2023dual} & 75.69 & 68.09 & 61.07 & 54.72 & 47.83 & 34.02 & 32.69 & 23.37 & 54.04 & 37.09 & 32.49 & 22.18 \\
GridMM~\cite{wang2023gridmm} & -- & -- & -- & -- & 51.37 & 36.47 & 34.57 & 24.56 & 53.13 & 36.60 & 34.87 & 23.45 \\ 
BEVBert~\cite{an2022bevbert} & 73.72 & 65.32 & 57.70 & 51.73 & 51.78 & 36.37 & 34.71 & 24.44 & 52.81 & 36.41 & 32.06 & 22.09 \\ \midrule
\textbf{GOAT (Ours)} & \textbf{78.64} & \textbf{71.40} & \textbf{63.74} & \textbf{57.85} & \textbf{53.37} & \textbf{36.70} & \textbf{38.43} & \textbf{26.09} & \textbf{57.72} & \textbf{40.53} & \textbf{38.32} & \textbf{26.70} \\ \bottomrule
\end{tabular}
\caption{Comparison with other state-of-the-art methods on the REVERIE dataset~\cite{qi2020reverie}. `--': unavailable statistics.}
\label{tab:reverie}
\end{table*}

\subsection{Comparisons with State-of-the-Arts}
\label{subsec:evaluation}
In~\cref{tab:r2r},~\ref{tab:reverie},~\ref{tab:rxr},~\ref{tab:soon}, we compare GOAT with the previous state-of-the-art (SoTA) methods on the R2R, REVERIE, RxR-English, and SOON datasets, respectively. On all these four datasets, our approach exhibits superior navigation performance, precise instruction-following alignment, and accurate object grounding across both seen and unseen environments. For instance, in R2R, GOAT achieves remarkable improvements in SPL compared to BEVBert~\cite{an2022bevbert}, with relative increases of 7.41\%, 6.45\%, and 4.74\% on three subsets. In REVERIE, GOAT shows substantial relative enhancements in RGSPL by 11.83\%, 6.55\%, and 20.87\% on three subsets. In the challenging SOON and RxR tasks, GOAT also exhibits significant improvements in performance metrics, highlighting its robustness and superior generalization capabilities over previous methods.

\begin{table}[]
\centering
\large
\setlength\tabcolsep{3pt}
\renewcommand{\arraystretch}{1}
\resizebox{\linewidth}{!}{
\begin{tabular}{@{}l|cccc|cccc@{}}
\toprule
\multirow{2}{*}{Method} & \multicolumn{4}{c|}{Validation Seen} & \multicolumn{4}{c}{Validation Unseen} \\
 & SR$\uparrow$ & SPL$\uparrow$ & nDTW$\uparrow$ & sDTW$\uparrow$ & SR$\uparrow$ & SPL$\uparrow$ & nDTW$\uparrow$ & sDTW$\uparrow$ \\ \midrule
Syntax~\cite{li2021improving} & 48.1 & 44.0 & 58.0 & 40.0 & 39.2 & 35.0 & 52.0 & 32.0 \\
SOAT~\cite{moudgil2021soat} & -- & -- & -- & -- & 44.2 & -- & 54.8 & 36.4 \\
HOP+~\cite{qiao2023hop_plus} & 53.6 & 47.9 & 59.0 & 43.0 & 45.7 & 38.4 & 52.0 & 36.0 \\
FOAM~\cite{dou2022foam} & -- & -- & -- & -- & 42.8 & 38.7 & 54.1 & 35.6 \\
ADAPT~\cite{lin2022adapt} & 50.3 & 44.6 & 56.3 & 40.6 & 46.9 & 40.2 & 54.1 & 37.7 \\
$\text{MAR}_{\text{M-MP}}$~\cite{kamath2023new} & -- & -- & -- & -- & 50.2 & -- & 60.3 & 43.9 \\ 
VLN-PETL~\cite{qiao2023vln} & 60.5 & 56.8 & 65.7 & 51.7 & 57.9 & 54.2 & 64.9 & 49.7 \\ \midrule
\textbf{GOAT (Ours)} & \textbf{74.1} & \textbf{68.1} & \textbf{71.0} & \textbf{61.4} & \textbf{68.2} & \textbf{61.7} & \textbf{67.1} & \textbf{56.6} \\ \bottomrule
\end{tabular}}
\caption{Comparison on the RxR-English dataset~\cite{ku2020room}.}
\label{tab:rxr}
\end{table}

\begin{table}[]
\centering
\large
\setlength\tabcolsep{3pt}
\renewcommand{\arraystretch}{1}
\resizebox{\linewidth}{!}{
\begin{tabular}{@{}l|cccc|cccc@{}}
\toprule
\multirow{2}{*}{Method} & \multicolumn{4}{c|}{Validation Unseen} & \multicolumn{4}{c}{Test Unseen} \\
 & OSR$\uparrow$ & SR$\uparrow$ & SPL$\uparrow$ & RGSPL$\uparrow$ & OSR$\uparrow$ & SR$\uparrow$ & SPL$\uparrow$ & RGSPL$\uparrow$ \\ \midrule
GBE~\cite{Zhu_2021_SOON} & 28.54 & 19.52 & 13.34 & 1.16 & 21.45 & 12.90 & 9.23 & 0.45 \\
DUET~\cite{chen2022think} & 50.91 & 36.28 & 22.58 & 3.75 & 43.00 & 33.44 & 21.42 & 4.17 \\
GridMM~\cite{wang2023gridmm} & 53.39 & 37.46 & 24.81 & 3.91 & 48.02 & 36.27 & 21.25 & 4.15 \\ \midrule
\textbf{GOAT (Ours)} & \textbf{54.69} & \textbf{40.35} & \textbf{28.05} & \textbf{5.85} & \textbf{50.63} & \textbf{40.50} & \textbf{25.18} & \textbf{6.10} \\ \bottomrule
\end{tabular}}
\caption{Comparison on the SOON dataset~\cite{Zhu_2021_SOON}.}
\label{tab:soon}
\end{table}

\subsection{Quantitative Analysis}
\label{subsec:ablation}
\noindent\textbf{1) Effect of Causal Inference.}
\begin{figure}
    \centering
    \includegraphics[width=1\linewidth]{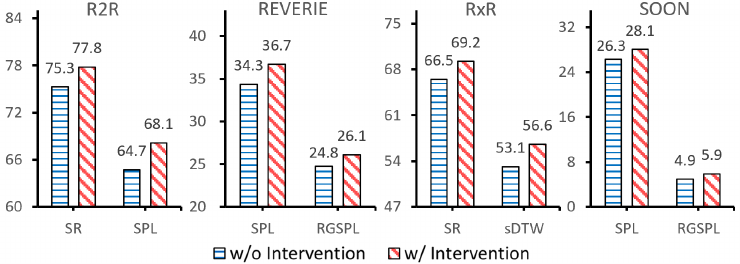}
    \caption{Effect of the intervention on various VLN datasets.}
    \label{fig:multi_dataset_ablation}
\end{figure}
\cref{fig:multi_dataset_ablation} verifies the impact of causal inference on GOAT across four diverse VLN datasets in unseen environments. ``W/o intervention" signifies the exclusion of the proposed BACL and FACL interventions across all modalities. On each of the datasets examined, the integration of causal inference leads to significant enhancements in the model's performance. This strongly demonstrates causal learning's considerable popularization potential in enhancing learning-based model generalization.

\noindent\textbf{2) Effect of BACL and FACL.}
\cref{tab:effect_causal_inference} analyzes the effects of the proposed BACL and FACL on the R2R val-unseen subset. Compared to the baseline (\#1), individual application of either BACL (\#2) or FACL (\#3) leads to performance improvements. Concurrent use of BACL and FACL (\#4) leads to further performance enhancements. These findings underscore our assumption about the presence of both observable and unobservable confounders. Integrating both back-door and front-door adjustments is crucial to comprehensively addressing dataset biases, and enhancing the model's robustness and generalization.

\noindent\textbf{3) Effect of CFP.}
\begin{table}[]
\centering
\small
\renewcommand{\arraystretch}{0.85}
\begin{tabular}{l|cc|cccc}
\toprule
Id & BACL & FACL & SR$\uparrow$ & SPL$\uparrow$ & NE$\downarrow$ & OSR$\uparrow$ \\ \midrule
1 & \ding{55} & \ding{55} & 75.27 & 64.69 & 2.72 & 83.14 \\
2 & \ding{51} & \ding{55} & 76.37 & 67.13 & 2.60 & 83.99 \\
3 & \ding{55} & \ding{51} & 76.50 & 66.92 & 2.53 & 84.21 \\
\textbf{4} & \ding{51} & \ding{51} & \textbf{77.82} & \textbf{68.13} & \textbf{2.40} & \textbf{84.72} \\ \bottomrule
\end{tabular}
\caption{Effect of back-door and front-door adjustments.}
\label{tab:effect_causal_inference}
\end{table}
\begin{table}[]
\centering
\small
\renewcommand{\arraystretch}{0.85}
\begin{tabular}{@{}c|l|l|cccc@{}}
\toprule
Stage & Id & Method & SR$\uparrow$ & SPL$\uparrow$ & NE$\downarrow$ & OSR$\uparrow$ \\ \midrule
\multirow{2}{*}{PT} & A0 & w/o CFP-P & 40.36 & 37.88 & 6.52 & 48.87 \\
 & A1 & w/ CFP-P & 43.47 & 40.90 & 6.06 & 53.13 \\ \midrule
\multirow{3}{*}{FT} & B0 & A0 w/o CFP-F & 75.56 & 65.90 & 2.63 & 82.42 \\
 & B1 & A1 w/o CFP-F & 76.63 & 66.17 & 2.63 & 84.67 \\
 & B2 & A1 w/ CFP-F & \textbf{77.82} & \textbf{68.13} & \textbf{2.40} & \textbf{84.72} \\ \bottomrule
\end{tabular}
\caption{Effect of CFP in pre-training and fine-tuning.}
\label{tab:effect_tim}
\end{table}
In~\cref{tab:effect_tim}, we assess the efficacy of the proposed CFP on the R2R val-unseen subset. During pre-training (PT), incorporating CFP as an additional auxiliary task (CFP-P) enhances training performance, improving SR and SPL by 3.11\% and 3.02\% (\#A1), respectively. In the fine-tuning (FT) stage, we compare the performance with and without the use of trained attention modules from CFP to extract global features for front-door dictionaries (CFP-F). ``W/o CFP-F" signifies the use of simple average pooling to compress features from the pre-trained model. \#B2 shows that the CFP provides more reliable confounder representations for causal learning (SPL $\uparrow$ 1.96\%).

\begin{table}[]
\centering
\small
\renewcommand{\arraystretch}{0.85}
\begin{tabular}{l|ll|cccc}
\toprule
Id & Text & Vision & SR$\uparrow$ & SPL$\uparrow$ & NE$\downarrow$ & OSR$\uparrow$ \\ \midrule
1 & Stats & Stats & 75.22 & 64.78 & 2.71 & 83.65 \\
2 & Stats & Attn & 76.59 & 65.39 & 2.56 & \textbf{85.31} \\
\textbf{3} & \textbf{Attn} & \textbf{Stats} & \textbf{77.82} & \textbf{68.13} & \textbf{2.40} & 84.72 \\
4 & Attn & Attn & 75.95 & 65.83 & 2.64 & 83.91 \\ \bottomrule
\end{tabular}
\caption{Effect of Statistic and Attention methods in BACL.}
\label{tab:effect_BACL}
\end{table}
\noindent\textbf{4) Effect of Different BACL in Different Modalities.}
\cref{tab:effect_BACL} investigates the effect of various combinations of statistic-based and attention-based methods for text and vision in BACL on the R2R val-unseen subset. The results indicate that employing the attention method for text and the statistic method for vision yields the best performance (\#3).
Intuitively, this can be explained by the structured nature of textual information and the involvement of RoBERTa's end-to-end training, enabling the attention method to effectively capture contextual nuances. Conversely, images lack explicit causality, and CLIP, the image extractor, isn't trained directly for efficiency reasons. Consequently, the statistic method ensures a stable causal learning process, preserving the integrity of vision-related features.

\subsection{Qualitative Analysis}
\label{subsec:visualization}
\begin{figure}
    \centering
    \includegraphics[width=1\linewidth]{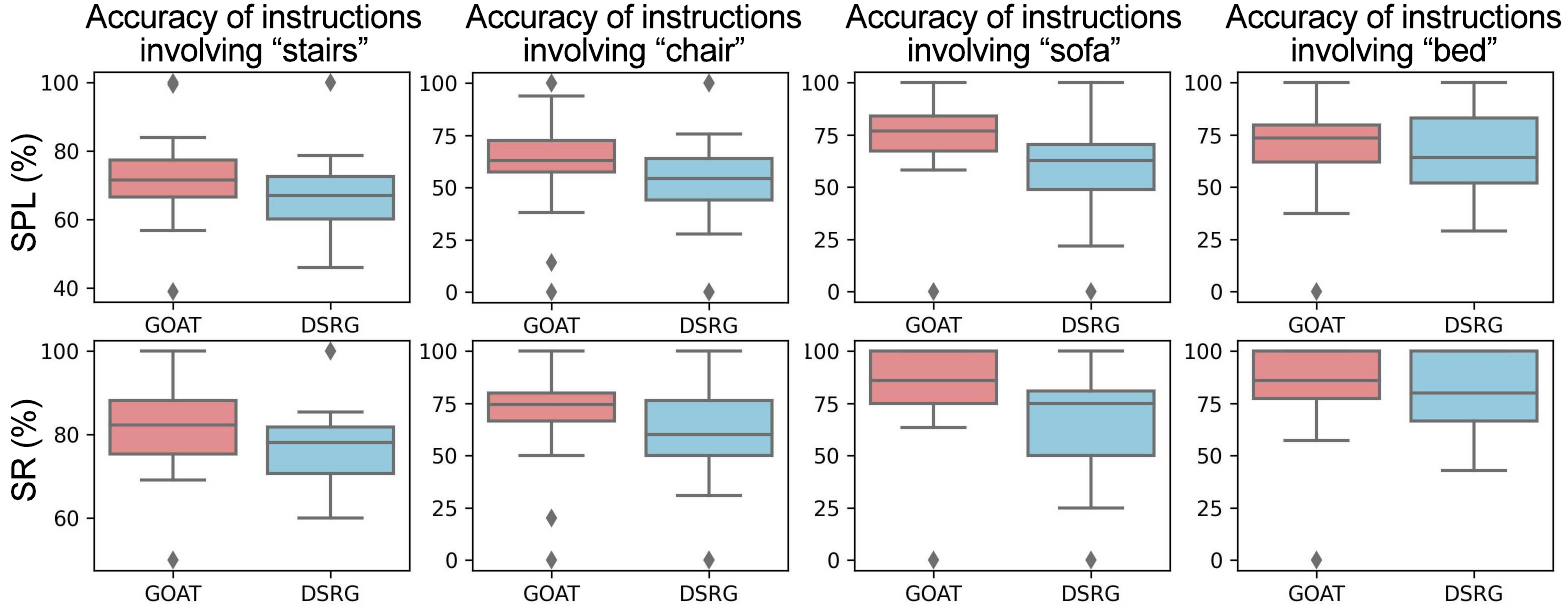}
    \caption{Comparison of the distribution of SR and SPL of instructions involving specific terms on the R2R val-unseen split.}
    \label{fig:vis_bias}
\end{figure}
\begin{figure}
    \centering
    \includegraphics[width=1\linewidth]{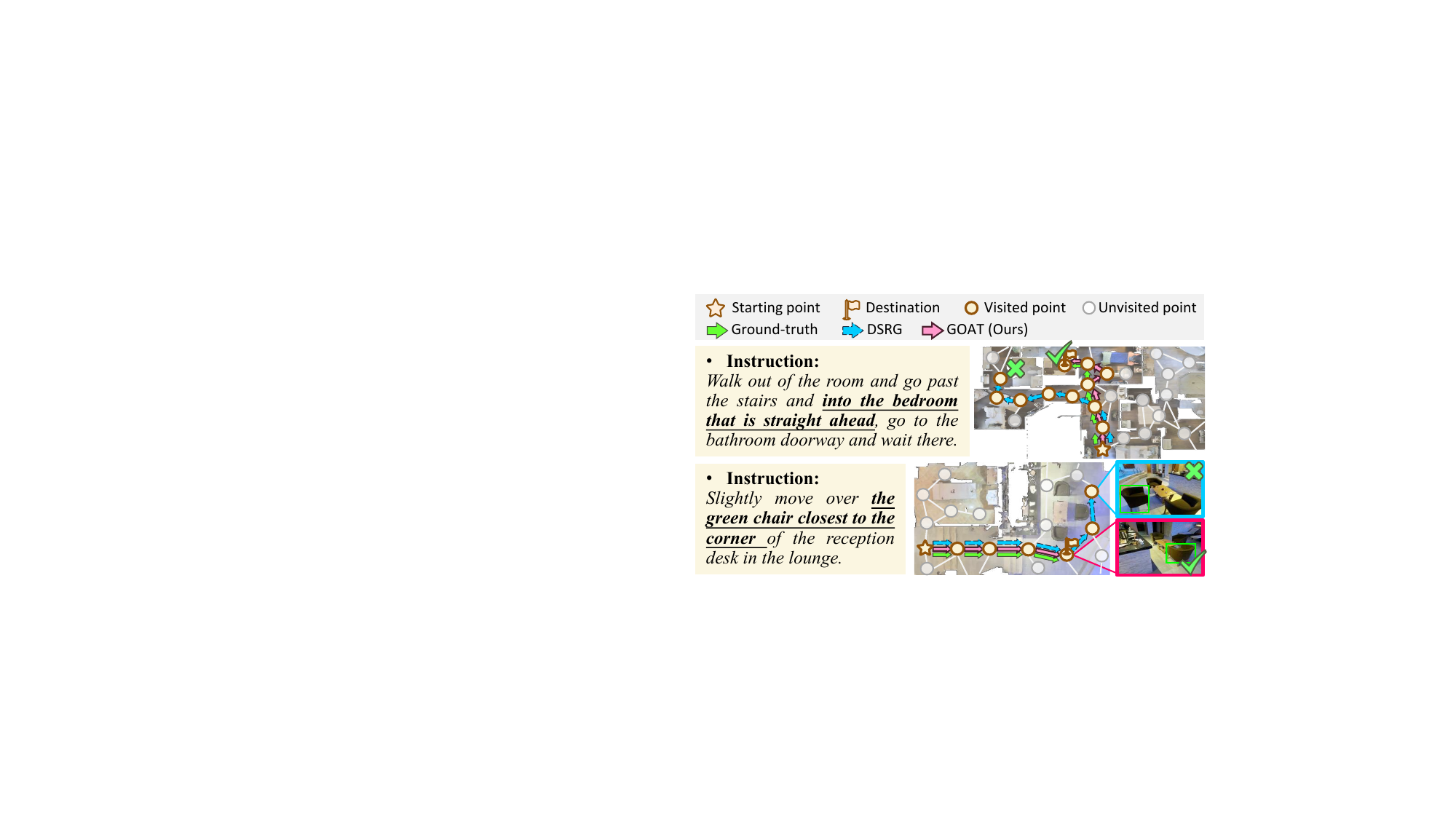}
    \caption{Predicted trajectories in unseen environments.}
    \label{fig:vis_path}
\end{figure}
\noindent\textbf{1) Bias Elimination Effect.} 
In~\cref{fig:vis_bias}, the compactness of the boxes represents concentrated data distribution and reduced variability, while a median line closer to the center signifies even data distribution. It shows that GOAT obtains narrower boxes and more central midlines across diverse objects. This finding showcases that with the integration of causal intervention, GOAT significantly reduces prediction bias, thereby enhancing its generalization capability in previously unseen environments.

\noindent\textbf{2) Visualized Trajectories.}
In~\cref{fig:vis_path}, we visualize some predicted trajectories in unseen environments, comparing them with DSRG on R2R and REVERIE datasets. Notably, GOAT precisely captures directional cues like ``straight ahead" and nuanced instructions like ``closest to the corner", enabling accurate predictions. These instances highlight the intricate causal connections in VLN tasks, where specific instructions prompt corresponding actions. GOAT's enhanced causal inference capability enables it to generate more reasoned responses aligned with the provided instructions, underscoring the significance of robust causal inference in VLN systems. Please refer to our supplementary material for more detailed discussions and visualizations.

\section{Conclusion}
\label{sec:conclusions}
Our work presents GOAT, a novel approach that addresses the dataset bias in VLN from the perspective of causal learning. The back-door and front-door adjustment causal learning (BACL and FACL) mechanisms are proposed to adjust for observable and unobservable confounders, respectively. The cross-modal feature pooling (CFP) module is adopted to promote feature learning and extraction through contrastive learning. The practical causal learning pipeline is presented to illuminate other similar learning-based methods. Experiments on R2R, REVERIE, RxR, and SOON datasets show that GOAT can reasonably discover sequence visual-linguistic causal structures and significantly improve performance. Beyond VLN, the underlying confounder assumption and causal inference principles are generalizable to other similar fields.

\section*{Acknowledgments}
This paper is supported by the National Natural Science Foundation of China under Grants (62233013, 62073245, 62173248). Shanghai Science and Technology Innovation Action Plan (22511104900).

{\small
\bibliographystyle{ieeenat_fullname}
\bibliography{main}
}

\clearpage \appendix
\section*{Appendix}
In this Appendix, we provide additional elaboration on aspects omitted in the main paper. 
\begin{itemize}
\setlength{\itemsep}{0pt}
\setlength{\parsep}{0pt}
\setlength{\parskip}{0pt}
\item \cref{sec: causal inference principles}: Elaborate derivation of back-door and front-door adjustments.
\item \cref{sec: dataset_appendix}: In-depth comparison of the four VLN datasets and corresponding metrics.
\item \cref{sec: experiment_appendix}: Additional experimental results and in-depth discussions on GOAT.
\item \cref{sec: failure_case}: Analysis of failure cases and comprehensive discussions of limitations.
\item \cref{sec: qualitative_appendix}: Additional qualitative panoramic visualizations from diverse datasets.
\end{itemize}

\section{Causal Inference Principles}
\label{sec: causal inference principles}
\subsection{Back-door Adjustment}
In the realm of causal inference~\cite{pearl2018book}, the back-door adjustment method serves as a cornerstone, enabling researchers to estimate causal effects from collected data. It hinges on understanding causality, allowing the assessment of the impact of an independent variable $X$ on a dependent variable $Y$ while minimizing the influence of confounders $Z$. It is essential to distinguish between \textit{``observation"} -- passive observation of natural relationships (typically formulated as $P(Y|X)=\sum_{z} P(Y|X,z)P(z|X)$) -- and \textit{``intervention"} -- active manipulation of variables to establish causality, denoted as $P(Y|\textit{do}(X))$. The \textit{do}-operator signifies an intervention where $X$ is forcibly set to a specific value $x$, thus blocking back-door causal paths originating from $X$.

To illustrate, consider $P(Y|X)$ and $P_m(Y|X)$ as probabilities before and after intervention on the causal graph, respectively, where $P(Y|do(X))=P_m(Y|X)$. Calculating the causal effect relies on the observation that $P_m$, the manipulated probability, shares two crucial properties with $P$. First, the marginal probability $P(Z=z)$ remains unchanged under intervention since the process determining $Z$ is not affected by removing the arrow from $Z$ to $X$, denoted as $P_m(z)=P(z)$. Second, the conditional probability $P(Y=y|X=x, Z=z)$ is invariant, because the process by which Y responds to X and Z remains consistent, regardless of whether X changes spontaneously or by deliberate manipulation, \ie, $P_m(Y|X,z)=P(Y|X,z)$. Additionally, the independence between $Z$ and $X$ under the intervention distribution leads to another rule: $P_m(z|X)=P_m(z)$. Considering these equations together, we can derive:
{\setlength\abovedisplayskip{1pt}
\setlength\belowdisplayskip{1pt}
\begin{align}
    P(Y|do(X)) &:= P_m(Y|X) \\
               &= \sum_{z} P_m(Y|X,z)P_m(z|X) \\
               &= \sum_{z} P_m(Y|X,z)P_m(z) \\
               &= \sum_{z} P(Y|X,z)P(z). \label{eq_backdoor_appendix}
\end{align}}
\cref{eq_backdoor_appendix} is called the \textit{back-door adjustment formula}. It computes the association between $X$ and $Y$ for each value $z$ of $Z$, then averages over those values. This procedure is referred to as ``adjusting for $Z$". This final expression can be estimated directly since it consists only of conditional probabilities. To better understand the concept of intervention and meanwhile demonstrate its effectiveness, we conducted a toy experiment based on~\cref{eq_toy_experiment1} and~\cref{eq_toy_experiment2} using direction-and-landmark keywords extracted from instructions in the R2R training dataset:
{\setlength\abovedisplayskip{1pt}
\setlength\belowdisplayskip{1pt}
\begin{align}
    P(Y|X) &= \frac{P(X,Y)}{P(X)} \label{eq_toy_experiment1} \\
    P(Y|do(X)) &= \sum_z \frac{P(Y,X,z)P(z)}{P(X,z)} \label{eq_toy_experiment2}
\end{align}}

\begin{figure}[tbp]
\centering
\subfloat[]{\includegraphics[width=0.5\linewidth]{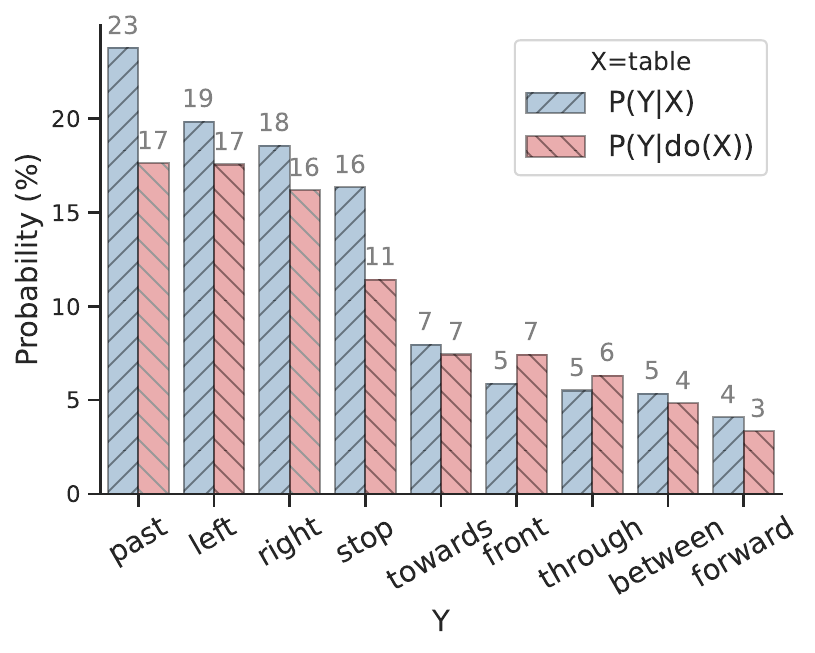}}
\subfloat[]{\includegraphics[width=0.5\linewidth]{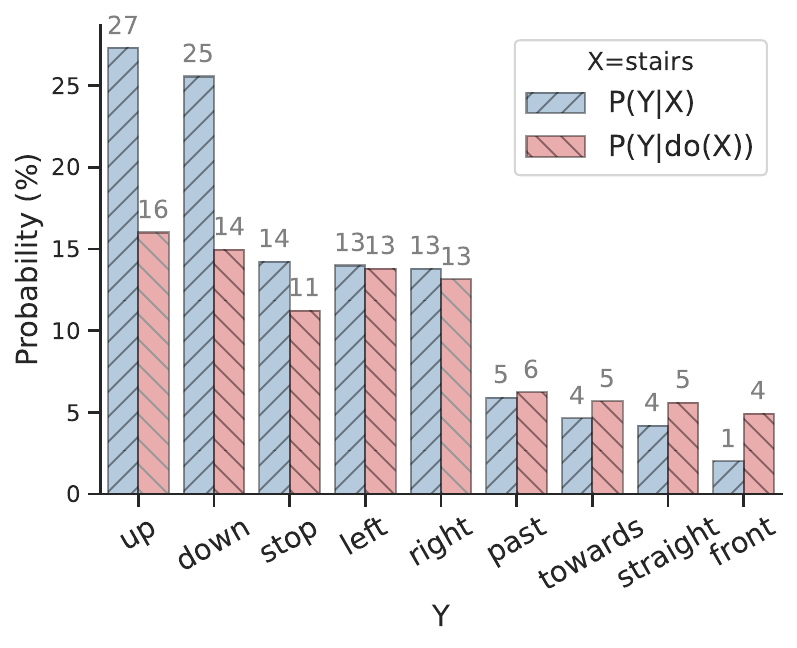}}
\caption{A toy experiment of the differences between the likelihood before (\ie, $P(Y|X)$) and after intervention (\ie, $P(Y|do(X))$) in the R2R training dataset. Only several cases are visualized to avoid clutter.}
\label{fig_vis_statistics}
\end{figure}

\begin{table*}[t]
\centering
\renewcommand{\arraystretch}{0.85}
\small
\begin{tabular}{l|l|cc|cc|cc|cc|c|c}
\toprule
\multirow{2}{*}{Task} & \multirow{2}{*}{Dataset} & \multicolumn{2}{c|}{Train} & \multicolumn{2}{c|}{Val Seen} & \multicolumn{2}{c|}{Val Unseen} & \multicolumn{2}{c|}{Test Unseen} & \multirow{2}{*}{\shortstack{Avg. \\Edge}} & \multicolumn{1}{l}{\multirow{2}{*}{\shortstack{Avg. \\Word}}} \\ \cline{3-10} 
 &  & Instr & House & Instr & House & Instr & House & Instr & House &  & \multicolumn{1}{l}{} \\ \midrule
\multirow{2}{*}{\shortstack{Fine-\\grained}} & R2R~\cite{anderson2018vision} & 14,039 & 61 & 1,021 & 56 & 2,349 & 11 & 4,173 & 18 & 5 & 29 \\
 & RxR-En~\cite{ku2020room} & 26,464 & 61 & 2,939 & 56 & 4,551 & 11 & 4,085 & 18 & 8 & 78 \\ \midrule
\multirow{2}{*}{\shortstack{Goal-\\oriented}} & REVERIE~\cite{qi2020reverie} & 10,466 & 60 & 1,423 & 46 & 3,521 & 10 & 6,292 & 16 & 5 & 18 \\
 & SOON~\cite{Zhu_2021_SOON} & 2,779 & 34 & 113 & 2 & 339 & 5 & 615 & 14 & 9 & 39 \\ \bottomrule
\end{tabular}
\caption{Dataset statistics. This table provides an overview of each split, including the number of instructions and houses, along with the average edge and average word count for each dataset.}
\label{tab: dataset statistic}
\end{table*}

As depicted in \cref{fig_vis_statistics}, it is evident that $P(Y|do(X))$ diverges from $P(Y|X)$, supporting our hypothesis that keywords within the instructions function as confounders. To illustrate, consider~\cref{fig_vis_statistics}(a) where $X$ represents a \texttt{table}. Previously biased probabilities associated with actions like \texttt{past}, \texttt{left}, and \texttt{right} become more balanced. In other words, when the agent encounters a table, its likelihood to move forward, left, and right becomes evenly distributed, mitigating the erroneous tendency introduced by dataset biases. Likewise, in~\cref{fig_vis_statistics}(b), the intricate probabilities surrounding \texttt{stairs} are unraveled, leading to a narrowing down of \texttt{up} and \texttt{down} probabilities to harmonize with other feasible actions like \texttt{stop}, \texttt{left}, and \texttt{right}. Therefore, by introducing the \textit{do}-operator to realize the active adjustment $P(Y|do(X))$ rather than merely passive observation $P(Y|X)$ during data fitting, the spurious correlations and underlying biases are alleviated.

\subsection{Front-door Adjustment}
While the back-door adjustment formula is effective to control for observable confounders, the front-door adjustment method steps in when the confounders cannot be directly observed. In essence, the front-door adjustment method tackles unobservable confounders by identifying alternative pathways that mediate the relationship between the input and the outcome. This nuanced approach is particularly valuable when dealing with intricate causal structures where certain variables are beyond direct measurement.

Concretely, an observable mediator $M$ is inserted between the input $X$ and the output $Y$, creating a front-door path $X \rightarrow M \rightarrow Y$. First, it's important to highlight that the influence of $X$ on $M$ can be identified, as there are no back-door paths from $X$ to $M$. Thus, we can obtain
{\setlength\abovedisplayskip{1pt}
\setlength\belowdisplayskip{1pt}
\begin{gather}
    P(M|do(X))=P(M|X). \label{eq:front_door_appendix_1}
\end{gather}}

Furthermore, it's crucial to recognize that the impact of $M$ on $Y$ is identifiable. This is because the back-door path from $M$ to $Y$ -- specifically, $M \leftarrow X \leftarrow Z \rightarrow Y$ -- can be blocked by conditioning on $X$:
{\setlength\abovedisplayskip{1pt}
\setlength\belowdisplayskip{1pt}
\begin{gather}
    P(Y|do(M))=\sum_{x'}P(Y|M,x')P(x') \label{eq:front_door_appendix_2}
\end{gather}}
where $x'$ denotes the possible value of the whole inputs, rather than the current input $X=x$. Both~\cref{eq:front_door_appendix_1} and~\cref{eq:front_door_appendix_2} are obtained through the adjustment formula. Subsequently, the front-door adjustment formula can be obtained by chaining these two partial effects:
{\setlength\abovedisplayskip{1pt}
\setlength\belowdisplayskip{1pt}
\begin{align}
    P(Y|do(X)) &=\sum_m(P(Y|do(M))P(M|do(X))) \label{eq:front_door_appendix_3} \\ 
    &= \sum_m\sum_{x'}P(Y|m,x')P(x')P(m|X). \label{eq:front_door_appendix_4}
\end{align}}

The integration of adjustment formulas, incorporating both the back-door and front-door criteria, encompasses diverse scenarios. By leveraging graphs and their underlying assumptions, we can more effectively discern causal relationships and derive causal representations from purely observational data. Motivated by the substantial potential of causal inference, this paper primarily focuses on approximating these adjustments for implementation in deep learning-based methods for VLN. To the best of our knowledge, this is the first work to explain VLN's hidden bias problem from the causal perspective and make an attempt to remove the effect caused by confounders via intervention.

\section{Datasets}
\label{sec: dataset_appendix}

\subsection{Comparison of Various VLN Datasets}
The statistical overview and comparison of the four VLN datasets are presented in \cref{tab: dataset statistic}.

\noindent\textbf{1. Fine-Grained VLN Datasets}, including R2R~\cite{anderson2018vision} and RxR~\cite{ku2020room}, offer detailed, step-by-step navigational instructions. Specifically, R2R is proposed to guide agents across rooms based on language instructions. RxR, an extension of R2R, augments the complexity with more intricate instructions and paths. To align with other VLN datasets, we focus on RxR's English subsets (en-IN and en-US). What sets the VLN challenge apart is the agent's necessity to follow varied language commands in previously \textit{unseen} real environments. This demands a high level of generalization capability, enabling adaptation to diverse situations.

\noindent\textbf{2. Goal-Oriented VLN Datasets} such as REVERIE~\cite{qi2020reverie} and SOON~\cite{Zhu_2021_SOON} emphasize object localization tasks, where agents must find specific objects based on remote referring descriptions. With additional object annotations, these goal-oriented datasets describe the target object and its location with concise instructions. 
The dataset splits of SOON outlined in DUET~\cite{chen2022think} are used for ensuring a unified evaluation approach. The goal-oriented VLN task enhances the high-level reasoning abilities of embodied agents, and provides valuable applications in real-world scenarios.

\subsection{Evaluation Metrics}
For fine-grained VLN tasks, the agent is expected to follow a specific path to reach the target location. The primary metric used to evaluate performance is the \texttt{Success Rate} (SR), indicating how often the agent completes the task within a certain distance (usually 3m) of the goal. Additionally, \texttt{Navigation Error} (NE) measures the average distance between the predicted and ground-truth locations. \texttt{Oracle Success Rate} (OSR) assesses whether any node in the predicted path is within a threshold of the target location. \texttt{Success weighted by Path Length} (SPL) is used to balance both success rate and trajectory length. Since RxR includes paths that approach the goal indirectly, two additional metrics are considered. \texttt{Normalized Dynamic Time Warping} (nDTW) penalizes deviations from the reference path to measure the match between two paths. \texttt{Success weighted by normalized Dynamic Time Warping} (sDTW) refines nDTW, focusing solely on successful episodes, thereby capturing both success and fidelity.
For goal-oriented tasks, the primary focus is on the agent's proximity to the goal. In addition to the above metrics, \texttt{Remote Grounding Success Rate} (RGS) is used to assess the accuracy of selecting the object from a set of candidates at the final position. \texttt{Remote Grounding Success Rate Weighted by Path Length} (RGSPL) is introduced to account for both success rate and path length.

\section{Additional Experimental Results}
\label{sec: experiment_appendix}

\begin{table}[]
\setlength\tabcolsep{3pt}
\centering
\small
\renewcommand{\arraystretch}{0.85}
\resizebox{0.8\linewidth}{!}{
\begin{tabular}{@{}l|l|cccc@{}}
\toprule
Id & Method & SR$\uparrow$ & SPL$\uparrow$ & NE$\downarrow$ & OSR$\uparrow$ \\ \midrule
\textbf{1} & \textbf{Full Model} & \textbf{77.82} & \textbf{68.13} & \textbf{2.40} & \textbf{84.72} \\ \midrule
2 & BACL w/o Text & 77.31 & 66.37 & 2.51 & 84.55 \\
3 & BACL w/o Vision & 75.95 & 65.31 & 2.65 & 83.78 \\ \midrule
4 & FACL w/o Text & 77.01 & 67.32 & 2.51 & 84.46 \\
5 & FACL w/o Vision & 76.97 & 67.07 & 2.51 & 83.23 \\
6 & FACL w/o History & 77.18 & 66.01 & 2.56 & 84.42 \\ \midrule
7 & Dict. w/o Update & 76.46 & 66.20 & 2.65 & 83.78 \\ 
8 & w/o AGF & 77.61 & 67.02 & 2.48 & 84.59 \\
9 & Inter. Only Final
& 76.29 & 66.80 & 2.58 & 84.55 \\ \bottomrule
\end{tabular}}
\caption{More ablation studies on the R2R val-unseen split.}
\label{tab: more_ablation}
\end{table}

\subsection{Confounder Factors Variation}
To explore the contribution of various modalities to causal learning, we further conducted detailed ablation studies on each modality. As demonstrated in \#2 -- \#6 in~\cref{tab: more_ablation}, different ablations lead to varying degrees of performance degradation, substantiating our hypothesis regarding confounders in VLN systems. Specifically, in BACL, the ablations of textual and visual intervention result in decreses in SR by 0.51\% and 1.81\%, and SPL by 1.76\% and 2.82\%, respectively. This suggests that visual intervention plays a crucial role, which is reasonable given that the primary distinction between seen and unseen environments lies in visual observation. In FACL, the ablations of textual, visual, and historical intervention lead to reductions in SR by 0.81\%, 0.85\%, and 0.64\%, and SPL by 0.81\%, 1.06\%, and 2.12\%, respectively. The adjustment to history has relatively more significant performance gains. Overall, these findings emphasize the importance of comprehensive interventions across cross-modal inputs, yielding more unbiased features and more generalized decision outcomes.

\subsection{Update of Confounder Dictionary}
In~\cref{tab: more_ablation} \#7, we investigate the impact of updating confounder dictionaries during training. This involves the textual confounder dictionary in BACL, supported by RoBERTa's end-to-end training, and random sampling from k-means clusters in FACL. The dictionaries are updated either when the model achieves a new best performance in the val-unseen split or every 3,000 iterations. The results demonstrate that updating the dictionary features aligns the representations of confounders more effectively with the evolving model weights and also enhances diversity, leading to improvements in overall performance ($\uparrow$ SPL 1.93\%).

\subsection{Adaptive Gate Fusion}
In~\cref{tab: more_ablation} \#8, we analyze the effects of the AGF module, designed to adaptively fuse causality-enhanced features and original context features using a gate-like structure. ``W/o AGF" signifies the direct use of causality-enhanced features without combining them with context features. The results demonstrate that the adaptive fusion process enables the model to effectively incorporate both types of features, leading to comparatively higher performance ($\uparrow$ SPL 1.11\%).

\subsection{Intervention Location}
In~\cref{tab: more_ablation} \#9, we validate the effectiveness of extending the assumption of causal learning to hidden features, rather than focusing solely on outputs. The results indicate that incorporating intervention modules only before the final Softmax layer enhances generalization capabilities to some extent. However, applying these interventions in shallower layers yields superior performance (SPL 68.13 \textit{vs.} 66.80). This extended assumption renders the application of causal learning in deep learning methods more flexible and practical.

\subsection{Number of K-Means Clusters in FACL}
\begin{figure}
    \centering
    \includegraphics[width=0.9\linewidth]{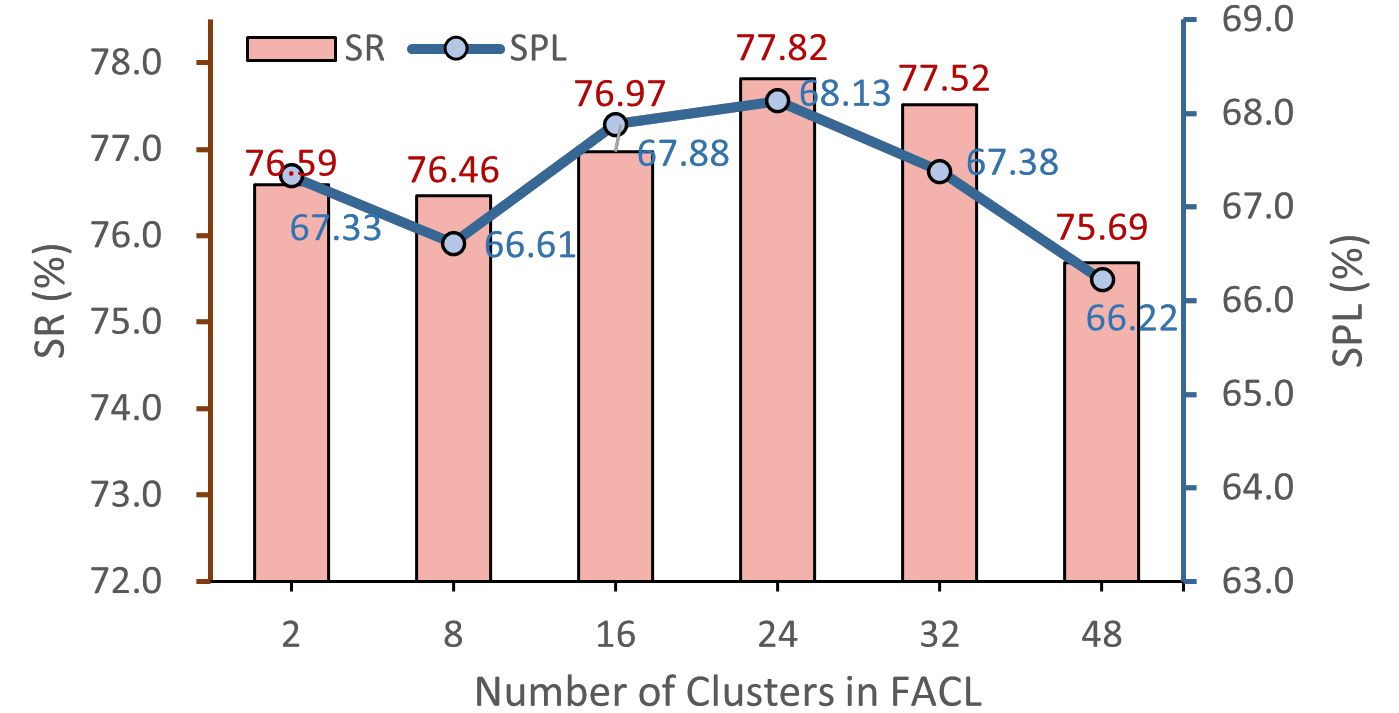}
    \caption{Effect of numbers of clusters in FACL.}
    \label{fig:number_cluster}
\end{figure}
Given that the confounder addressed by FACL is unobservable, we employ the K-Means algorithm to cluster the global features extracted by the trained CFP module from the entire training dataset. During the fine-tuning phase, we periodically sample features from these clusters to integrate into training. The experimental results of determining the number of categories for clusters are shown in~\cref{fig:number_cluster}. It presents that the choice of 24 clusters in FACL yields optimal performance in both SR and SPL. This strategic clustering approach ensures a comprehensive coverage of potential categories, enhancing the model's ability to discern confounders. Notably, too few clustering categories may overlook crucial distinctions, whereas an excessive number introduces redundant computational overhead and irrelevant noise, ultimately hampering training performance.

\subsection{Efficiency and Effectiveness Comparison}
\begin{figure}
    \centering
    \includegraphics[width=1.0\linewidth]{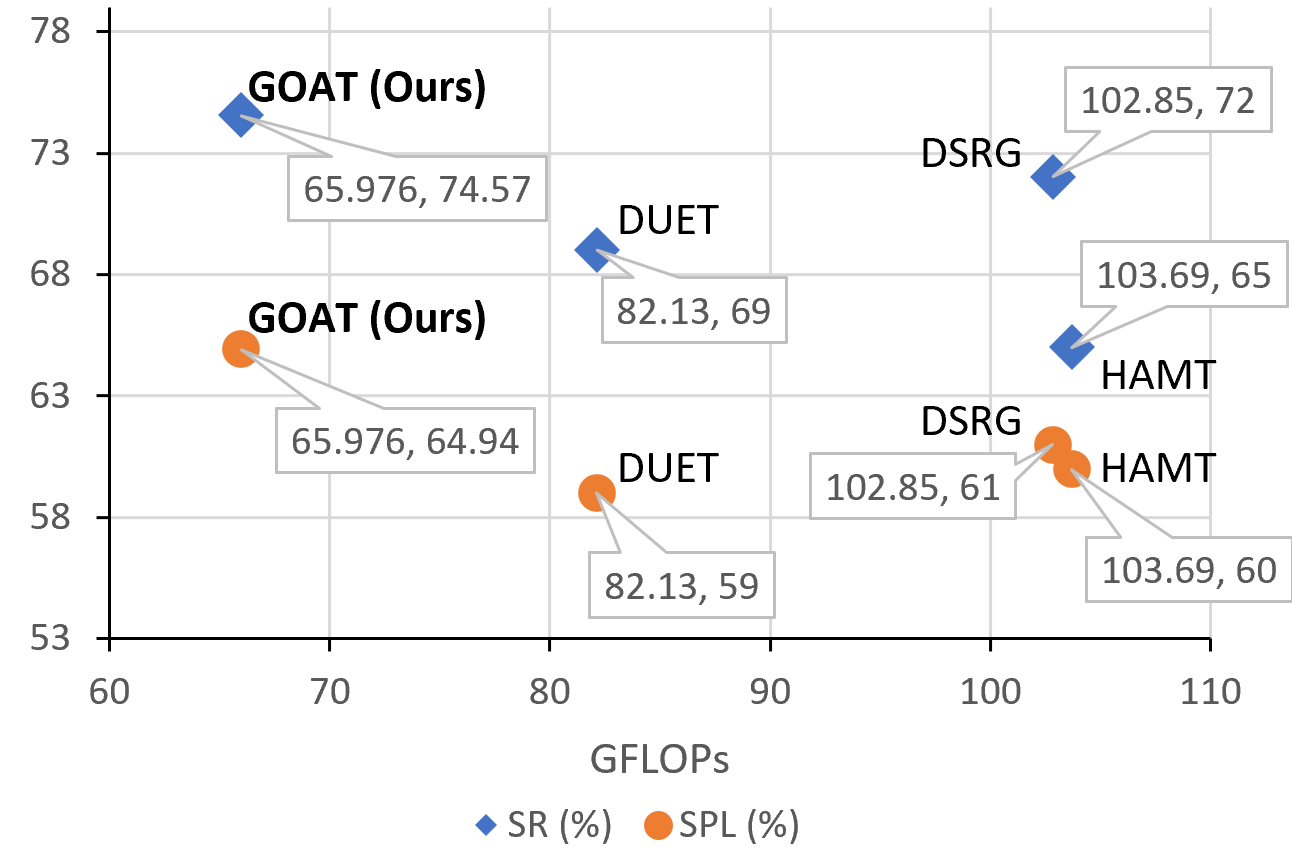}
    \caption{Comparison of GFLOPs and Accuracy.}
    \label{fig:GFLOPs}
\end{figure}
It is necessary to consider both efficiency and effectiveness since VLN is prompted to be applicable in real-world robots in the future. To assess computational complexity, we employed the Python toolkit \texttt{thop}, comparing GFLOPs with other transformer-based methods. For fair comparison, we conducted single-step forward inference with a batch size of 8, instruction length of 44, and historical global graph node of 6 across all methods. As shown in~\cref{fig:GFLOPs}, GOAT strikingly balances efficiency and effectiveness, outperforming previous approaches in both SR and SPL while maintaining lower GFLOPs. This reduction in computational cost is attributed to the adoption of a lighter framework with fewer transformer layers. This discovery illustrates that in scenarios with restricted task-specific datasets, adopting a lighter framework can enhance generalization while significantly reducing computational costs.

\section{Analysis of Failure Cases and Limitations}
\label{sec: failure_case}
\begin{figure}
    \centering
    \includegraphics[width=1.0\linewidth]{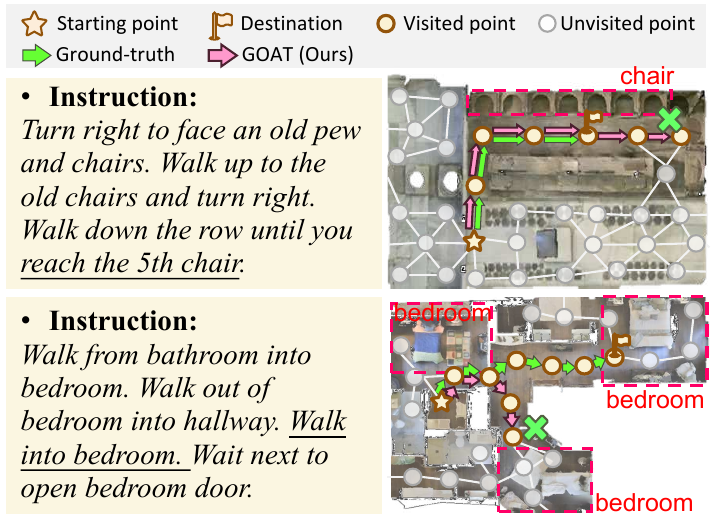}
    \caption{Illustration of Failure Cases.}
    \label{fig:failure_case}
\end{figure}
Despite GOAT's remarkable performance, we also examined specific failure cases to shed light on its limitations. For instance, as depicted in~\cref{fig:failure_case}, GOAT struggles with instructions involving numerical references. In the first case, it misidentified the \texttt{5th} chair but arrived at the \texttt{8th} chair instead. This phenomenon is aligned with the problem of current large models that are not sensitive to numbers and arithmetic tasks. Addressing this issue could benefit from approaches like the chain-of-thought method~\cite{wei2023chainofthought}, which has shown promise in handling numerical tasks. Moreover, when instructions are initially ambiguous (\eg, in the second case, there are actually \texttt{two bedrooms} that fit the description), GOAT might select the wrong option. Utilizing datasets like~\cite{nguyen2019help,thomason2020vision}, which focus on human-agent interaction, could improve the agent's decision-making in response to ambiguous instructions. Incorporating such datasets could empower the agent with a more robust and practical interactive capacity, reducing the likelihood of erroneous predictions. 
Finally, the limitation to the integration of causal learning with deep-learning methods, including the approximation process inherent in calculating expectations, and distinct modalities exhibiting varying preferences for specific probability estimations, requires ongoing efforts in future research to enhance the interpretability of these discrepancies.

\section{Additional Qualitative Examples}
\label{sec: qualitative_appendix}
\begin{figure*}[t]
    \centering
    \begin{minipage}{0.85\linewidth}
        \centering
        \includegraphics[width=1\linewidth]{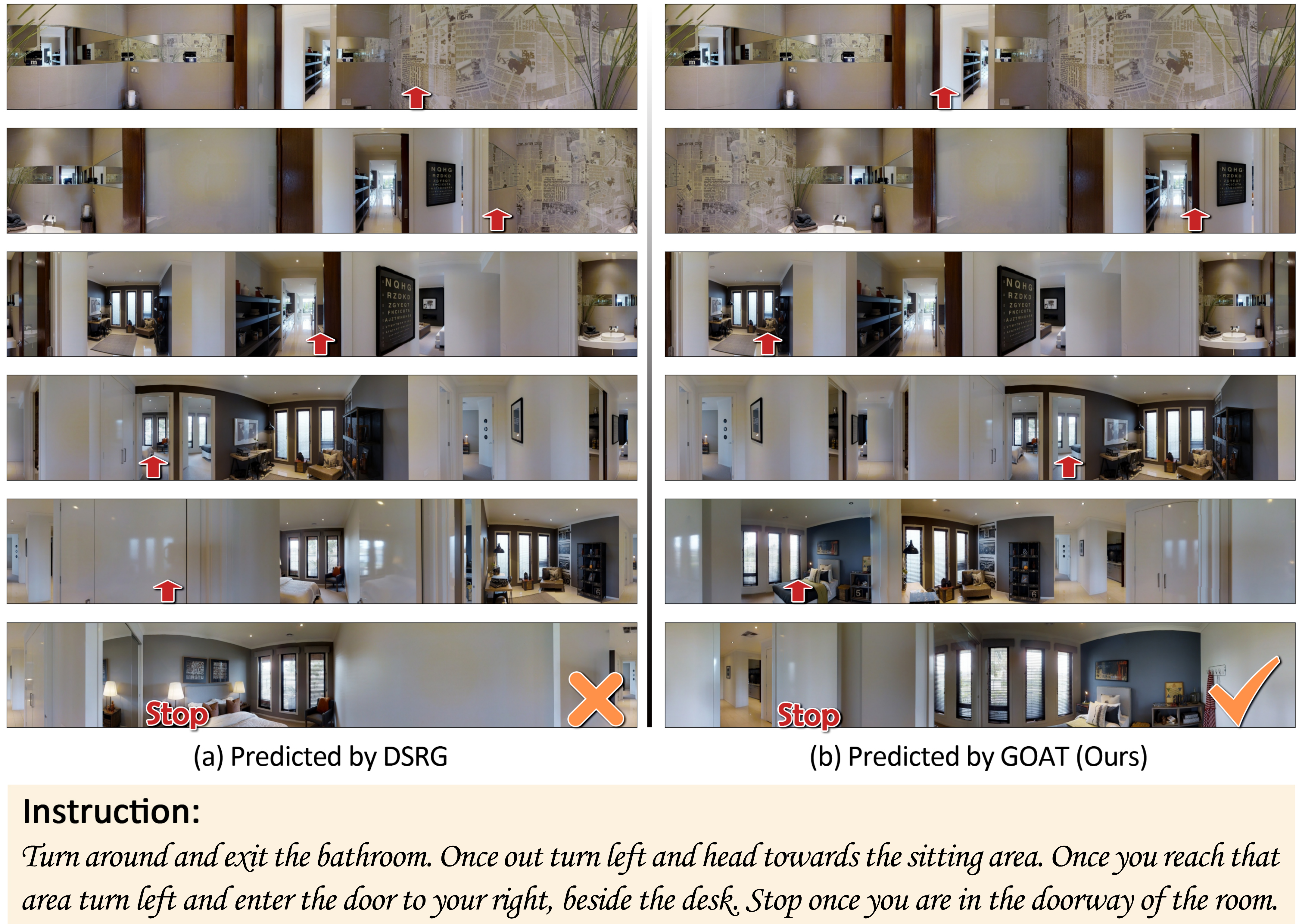}
        \vspace{3mm}
    \end{minipage}
    
    \begin{minipage}{0.85\linewidth}
    \centering
    \includegraphics[width=1\linewidth]{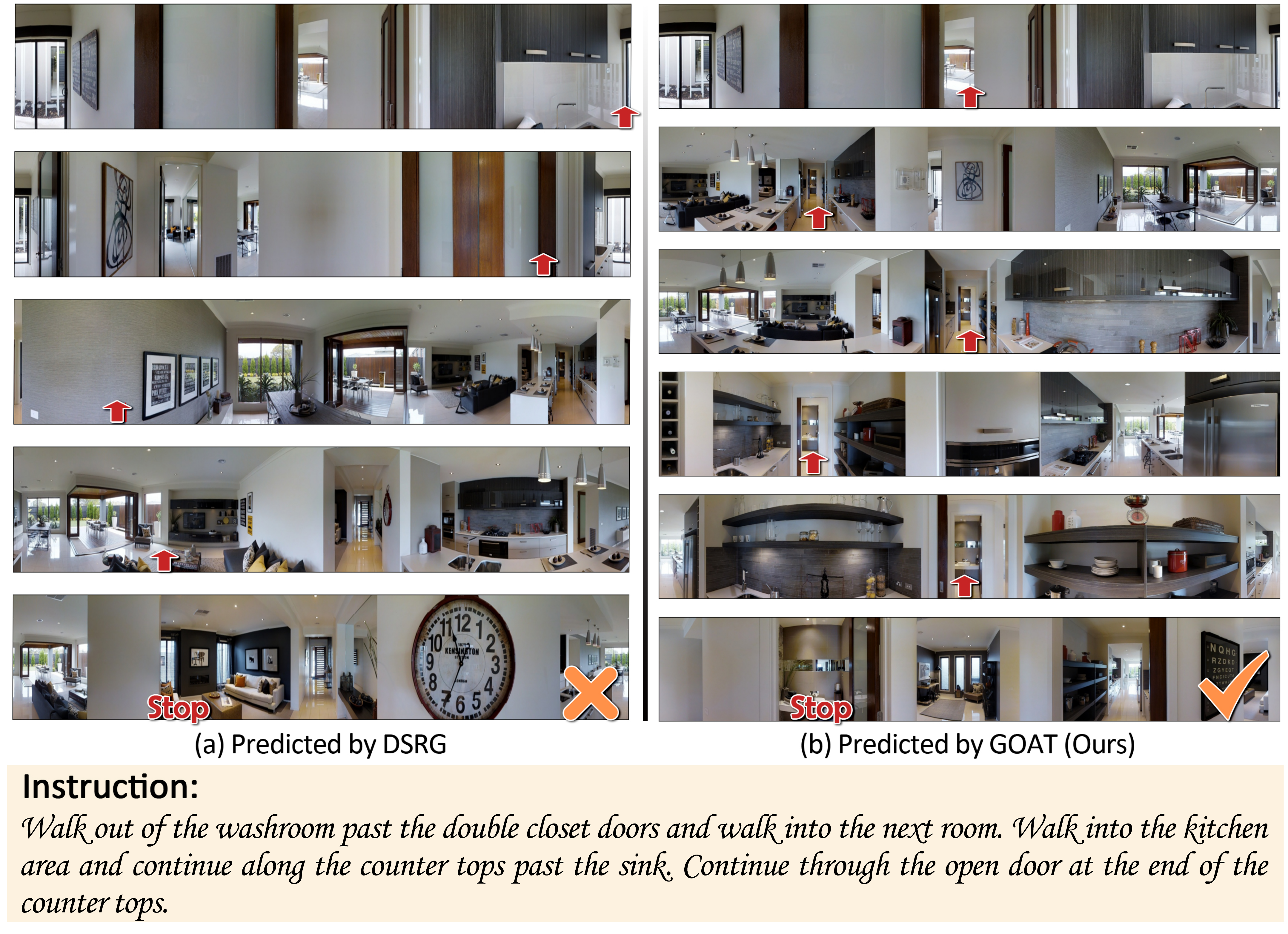}
    \end{minipage}
    
    \caption{Visual examples in the R2R validation-unseen split with navigation instructions presented at the bottom.}
    \label{fig: pano_r2r_1}
\end{figure*}

\begin{figure*}[t]
    \centering
    \begin{minipage}{0.85\linewidth}
        \centering
        \includegraphics[width=1\linewidth]{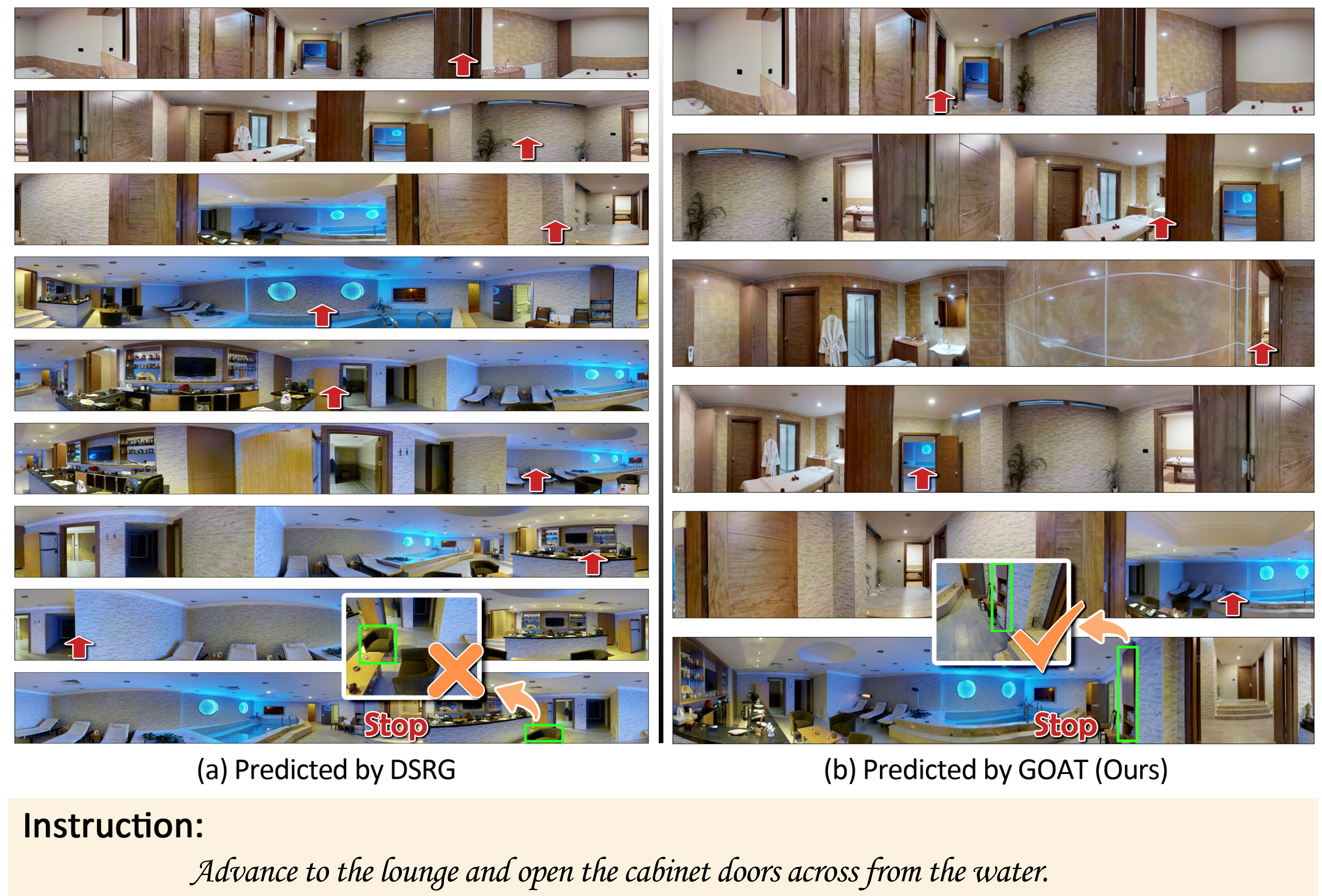}
        \vspace{3mm}
    \end{minipage}
    
    \begin{minipage}{0.85\linewidth}
    \centering
    \includegraphics[width=1\linewidth]{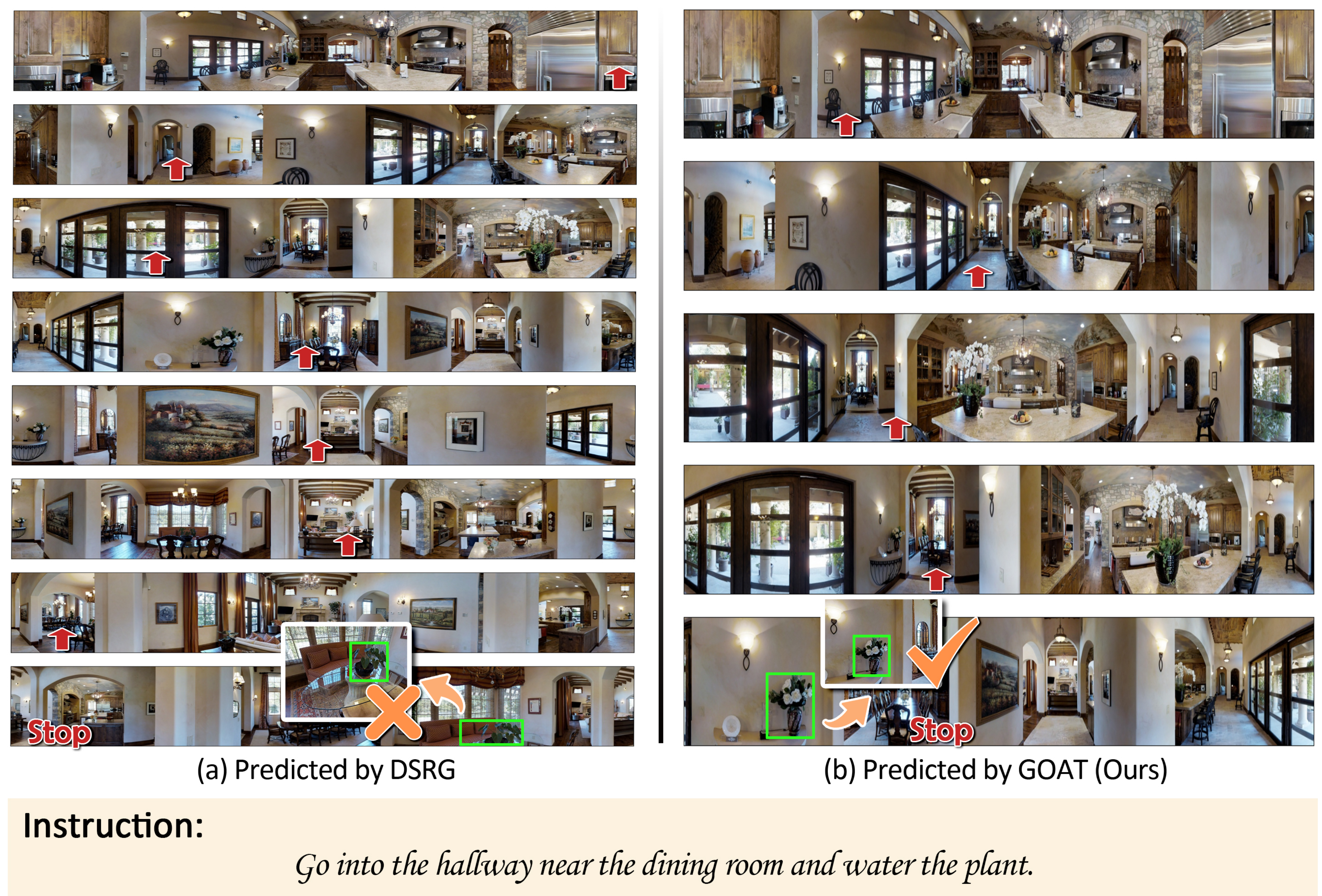}
    \end{minipage}
    
    \caption{Visual examples in the REVERIE validation-unseen split with navigation instructions presented at the bottom.}
    \label{fig: pano_reverie}
\end{figure*}

\begin{figure*}[t]
    \centering
    \begin{minipage}{0.85\linewidth}
        \centering
        \includegraphics[width=1\linewidth]{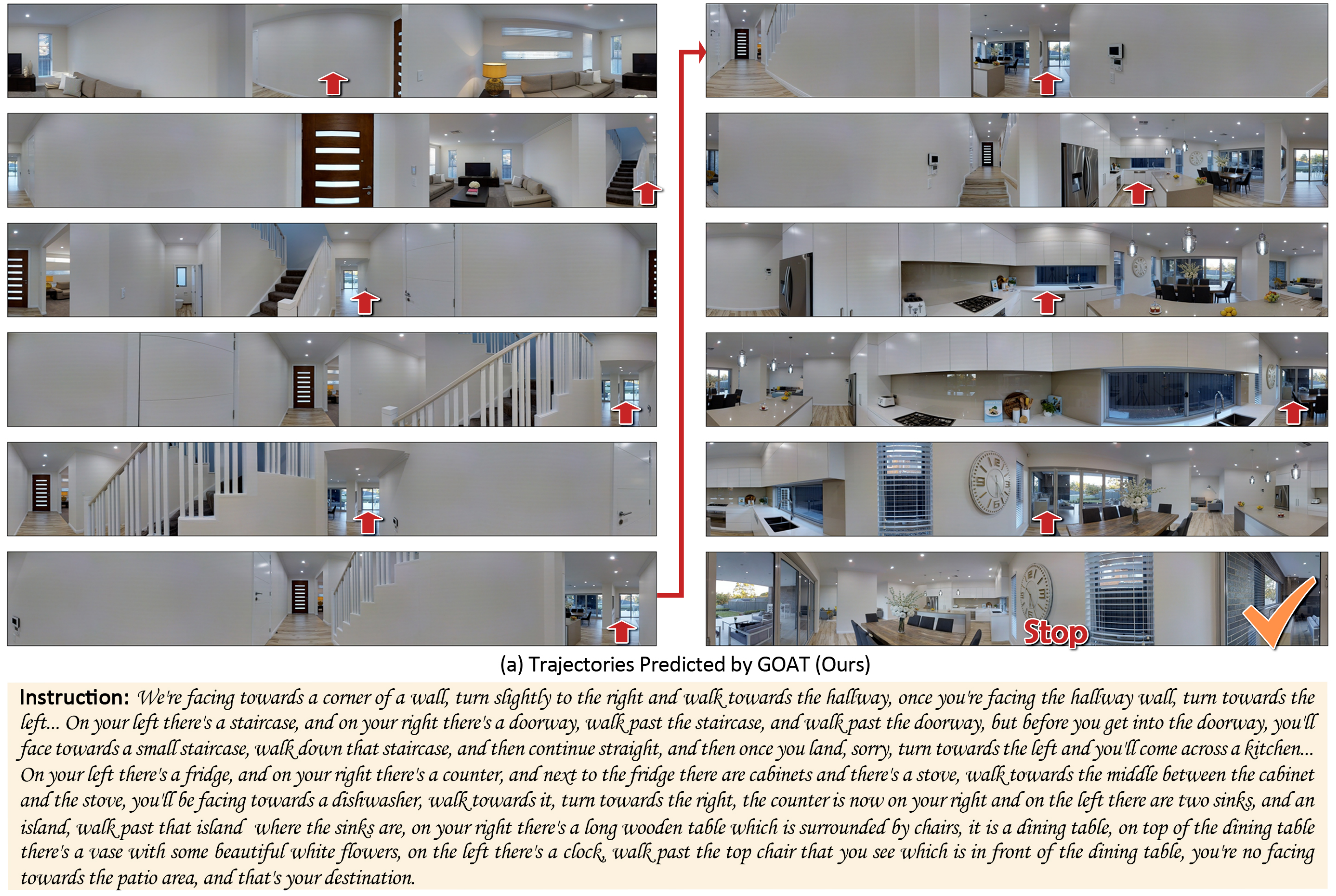}
        \vspace{3mm}
    \end{minipage}
    
    \begin{minipage}{0.85\linewidth}
    \centering
    \includegraphics[width=1\linewidth]{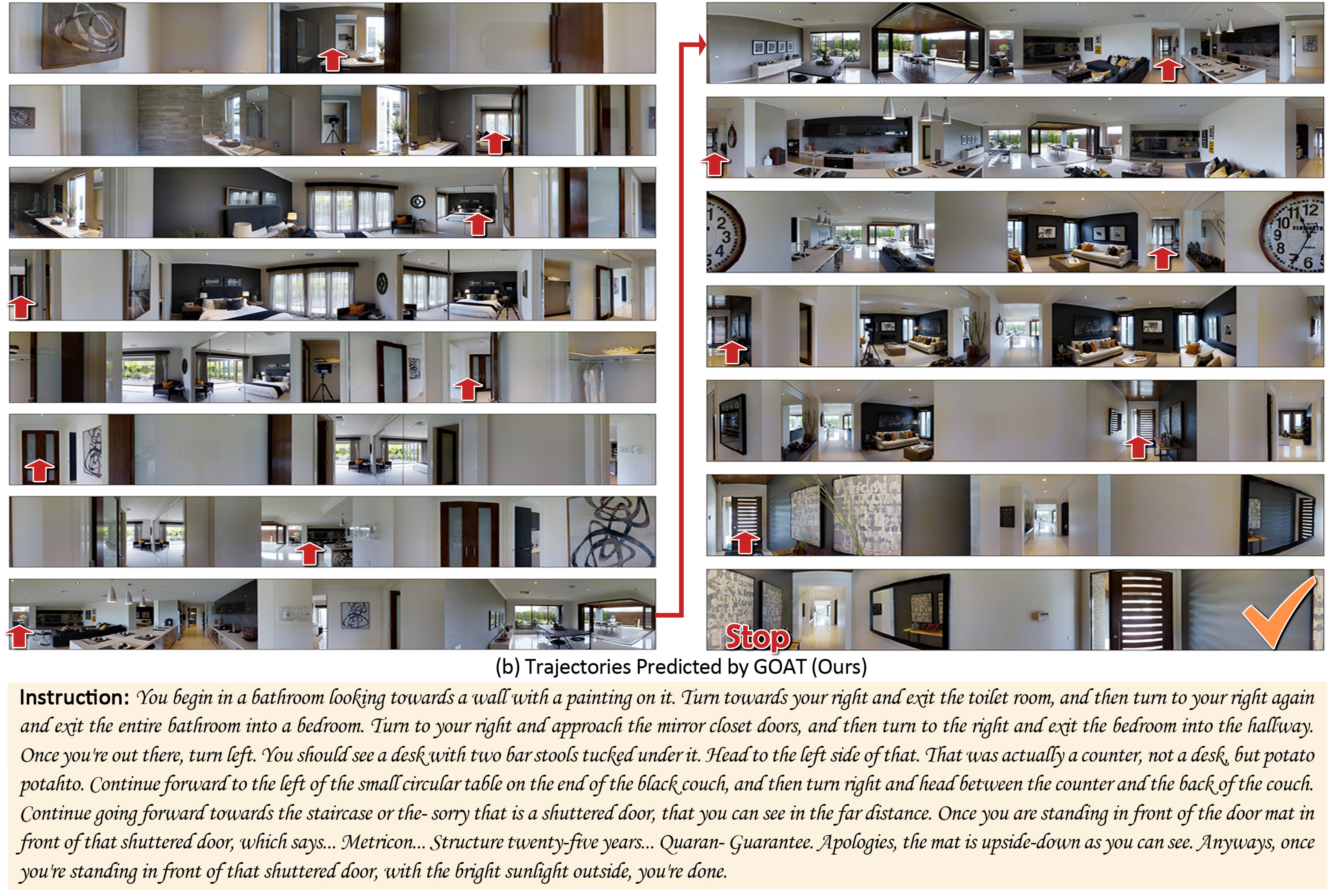}
    \end{minipage}
    
    \caption{Visual examples in the RxR validation-unseen split with navigation instructions presented at the bottom.}
    \label{fig: pano_rxr}
\end{figure*}

\begin{figure*}[t]
    \centering
    \begin{minipage}{0.85\linewidth}
        \centering
        \includegraphics[width=1\linewidth]{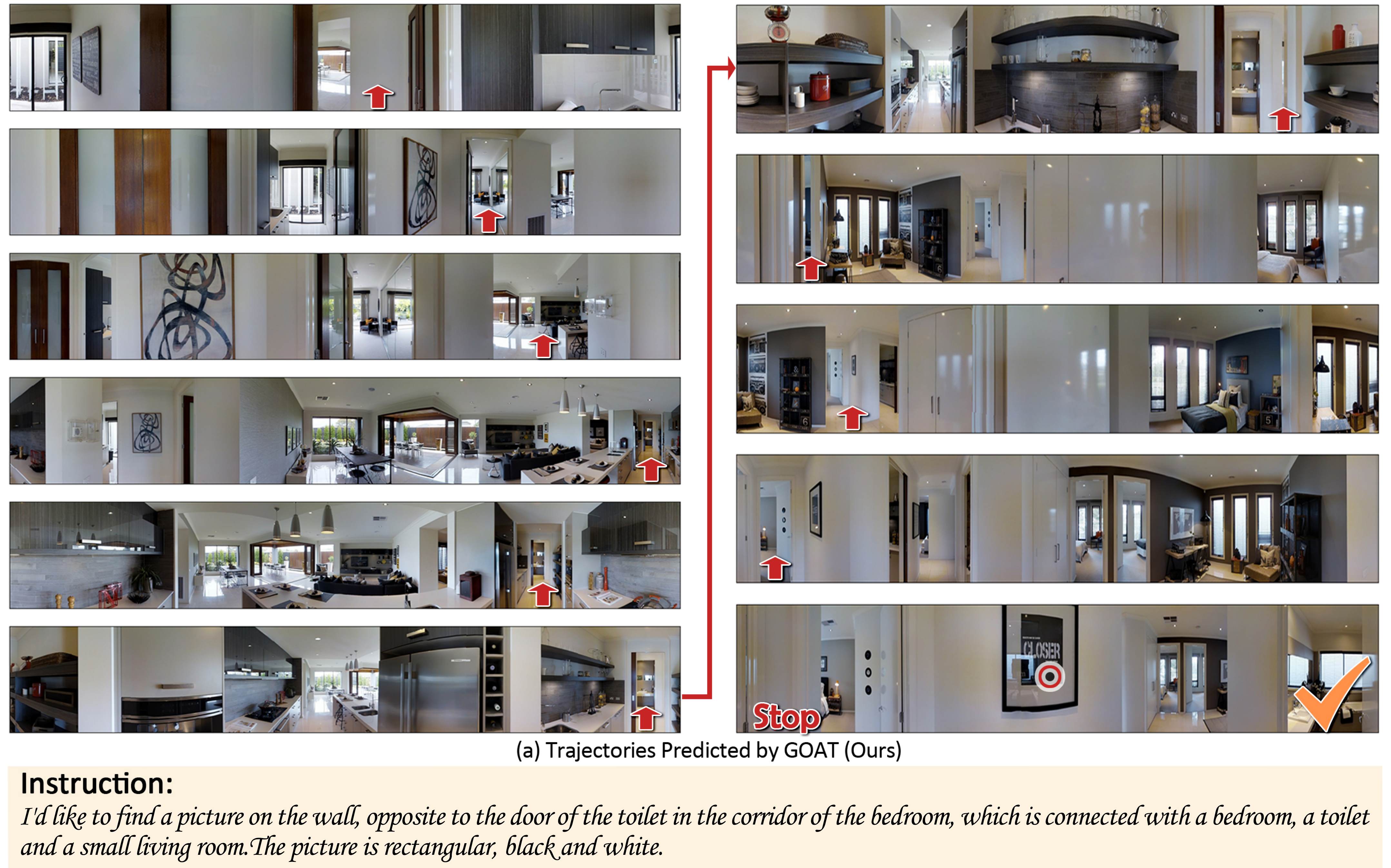}
        \vspace{3mm}
    \end{minipage}
    
    \begin{minipage}{0.85\linewidth}
    \centering
    \includegraphics[width=1\linewidth]{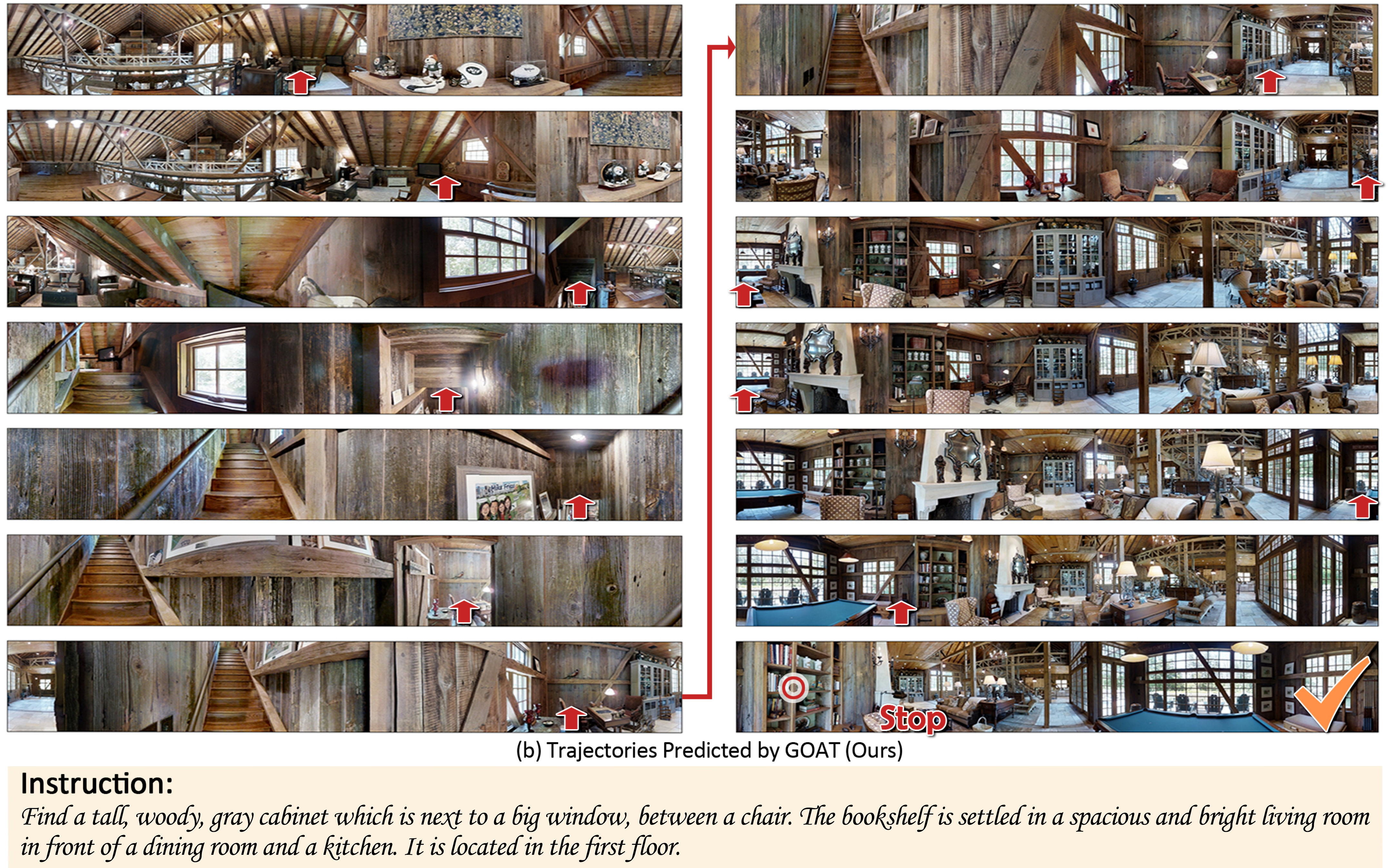}
    \end{minipage}
    
    \caption{Visual examples in the SOON validation-unseen split with navigation instructions presented at the bottom.}
    \label{fig: pano_soon}
\end{figure*}

Due to space limitations in the text, we present only the top view to depict the navigation scenarios. In this section, we provide predicted panoramic paths on four datasets in~\cref{fig: pano_r2r_1}, \ref{fig: pano_reverie}, \ref{fig: pano_rxr}, \ref{fig: pano_soon}, respectively, to enhance readers' comprehension of the tasks' objectives and the effectiveness of our approach. Specifically, the red arrows indicate the forward directions, and the corresponding instructions are provided below the visualized trajectories.

\end{document}